\documentclass{article} % For LaTeX2e
\usepackage{iclr2026_conference,times}

\usepackage{amsmath,amsfonts,bm}

\def\eqref#1{equation~\ref{#1}}
\def\1{\bm{1}}

\DeclareMathAlphabet{\mathsfit}{\encodingdefault}{\sfdefault}{m}{sl}
\SetMathAlphabet{\mathsfit}{bold}{\encodingdefault}{\sfdefault}{bx}{n}

\usepackage[utf8]{inputenc} % allow utf-8 input
\usepackage[T1]{fontenc}    % use 8-bit T1 fonts
\usepackage{hyperref}       % hyperlinks
\usepackage{url}            % simple URL typesetting
\usepackage{booktabs}       % professional-quality tables
\usepackage{amsfonts}       % blackboard math symbols
\usepackage{nicefrac}       % compact symbols for 1/2, etc.
\usepackage{microtype}      % microtypography
\usepackage{xcolor}         % colors
\usepackage{multirow} % 用于合并单元格
\usepackage{array} % 用于调整列格式
\usepackage{makecell}
\usepackage{varwidth} 
\usepackage{booktabs}
\usepackage{multirow}
\usepackage{microtype}      % microtypography
\usepackage{enumitem}
\usepackage{amsmath}
\usepackage{amssymb}
\usepackage{multirow}
\usepackage{color}
\usepackage{xcolor}
\usepackage{colortbl}
\usepackage{subcaption}
\usepackage{pifont}
\usepackage{bbding}
\usepackage[most]{tcolorbox} 
\newcommand{\ours}{{Vlaser}}

\usepackage{CJK}
\makeatletter
\def\blfootnote{\xdef\@thefnmark{}\@footnotetext}
\makeatother

\title{Vlaser: Vision-Language-Action Model with Synergistic Embodied Reasoning}

\author{
    \normalsize{}
    \textbf{Ganlin Yang$^{1,2*}$,~Tianyi Zhang$^{4,2*}$,~Haoran Hao$^{5,2*}$,~Weiyun Wang$^{6,2}$, 
    ~Yibin Liu$^{9,3}$, Dehui Wang$^{3}$}
    \\
    \normalsize{}\textbf{
    ~Guanzhou Chen$^{3,2}$, ~Zijian Cai$^{10,3}$, ~Junting Chen$^{8,2}$, ~Weijie Su$^{2}$,  ~Wengang Zhou$^1$, ~Yu Qiao$^2$} \\
    \normalsize{}
   \textbf{~Jifeng Dai$^{7,2}$, ~Jiangmiao Pang$^2$, ~Gen Luo$^{2}$,
    ~Wenhai Wang$^2$, ~Yao Mu$^{3,2\dag}$, ~Zhi Hou$^{2\dag}$}  \\
     [1mm]
     \normalsize{}
    ~$^1$University of Science and Technology of China \quad $^2$Shanghai AI Laboratory 
    \\ ~~\normalsize{}$^3$Shanghai Jiao Tong University  ~~\normalsize{}$^4$Zhejiang University ~~\normalsize{}$^5$Nanjing University ~~\normalsize{}$^6$Fudan University\\ ~~\normalsize{} $^7$Tsinghua University  ~~\normalsize{}  $^8$NUS ~~\normalsize{} $^9$Northeastern University ~~\normalsize{} $^{10}$Shenzhen University\\
    [3mm]
    \normalsize{}
    Project Page: 
    \href{https://internvl.github.io/blog/2025-10-11-Vlaser/}{Vlaser}  
}
\date{}

\iclrfinalcopy
\begin{document}

\maketitle
\blfootnote{\noindent$^{*}$Equal contribution. $^{\dag}$Corresponding authors.}

\begin{abstract}

While significant research has focused on developing embodied reasoning capabilities using Vision-Language Models (VLMs) or integrating advanced VLMs into Vision-Language-Action (VLA) models for end-to-end robot control, few studies directly address the critical gap between upstream VLM-based reasoning and downstream VLA policy learning. In this work, we take an initial step toward bridging embodied reasoning with VLA policy learning by introducing \textbf{\ours} -- a \textbf{V}ision-\textbf{L}anguage-\textbf{A}ction Model with \textbf{s}ynergistic \textbf{e}mbodied \textbf{r}easoning capability, which is a foundational vision-language model designed to integrate high-level reasoning with low-level control for embodied agents. Built upon the high-quality~\ours-6M dataset,~\ours~achieves state-of-the-art performance across a range of embodied reasoning benchmarks—including spatial reasoning, embodied grounding, embodied QA, and task planning.
Furthermore, we systematically examine how different VLM initializations affect supervised VLA fine-tuning, offering novel insights into mitigating the domain shift between internet-scale pre-training data and embodied-specific policy learning data. Based on these insights, our approach achieves state-of-the-art results on the WidowX benchmark and competitive performance on the Google Robot benchmark. The code, model and data are available at \url{https://github.com/OpenGVLab/Vlaser/}.

% —and further sets a new state of the art on the real-to-sim benchmark SIMPLER by finetuning on the manipulation datasets. Building on this architecture, we systematically investigate the impact of different vision-language models on supervised VLA fine-tuning, and provide novel insights for future foundational embodied vision-language construction: it is urgent to shrink the domain gap between internet data and the embodied scenes. 

% In this work, we present \ours, a foundational vision-language-action model that integrates embodied reasoning with robot control for embodied agent. \ours~achieves state-of-the-art performance on diverse embodied reasoning benchmarks—including spatial reasoning, point grounding, embodied QA, and robot planning—and further sets a new state of the art on the real-to-sim benchmark SIMPLER by finetuning on the manipulation datasets. Building on this architecture, we systematically investigate the impact of different vision-language models on supervised VLA fine-tuning, and provide novel insights for future foundational embodied vision-language construction: it is urgent to shrink the domain gap between internet data and the embodied scenes. 

% Model is open-source

\end{abstract}

\section{Introduction}

% 什么意义，当前状况，为啥我们要做这个事情，有啥意义。

% 为啥要端到端

% Embodied artificial intelligence is xxx, perception, understanding, action, 

% embodied reasoning & vision-language-action, 
% point out the value.

% Intuitively, vision-language models are capable of improving the perception generalization and model reasoning. 
% Actually, vision-language models with current spatial xx, is not benefit due to the domain shift.

% VLA requires in-domain finetuing.

% in domain data xxx, provide new insight for constructing vision-language-action foundation models.

% The generalization of policy is far from availability

% include navigation

% system1, system2.

% 当前的问题:
% 1. 没有拉齐，哪部分多模态标注真管用
% 2. 忽略了传统方法的优势，VLM. VLM 能够提供推理能力，改善泛化能力。

% -> 端到端？

Embodied artificial intelligence (AI)~\citep{chrisley2003embodied} aims to endow agents with the ability to perceive, understand, and act in the physical world. Achieving such intelligence requires not only accurate perception and language understanding but also embodied reasoning and effective control, which together define the paradigm of vision-language-action (VLA) models. Developing foundation models that possess strong reasoning and control capabilities is therefore an important advancement toward general-purpose embodied AI.

% In this context, 

In this context, vision-language models (VLMs)~\citep{openai2023gpt4v,liu2023visual,chen2024internvl,bai2025qwen2, team2023gemini} emerge as natural candidates to enhance embodied agents in perception generalization and reasoning ability. Following this paradigm, extensive embodied vision-language models~\citep{azzolini2025cosmos,team2025gemini} emerge from enhancing the key ability for an embodied agent in grounding, planning, and spatial reasoning. 
Meanwhile, a significant body of work extends vision-language models (VLMs) into vision-language-action models (VLAs)~\citep{kim2024openvla,intelligence2025pi_,driess2025knowledge} for robot control.
While there are some approaches~\citep{intelligence2025pi_,driess2025knowledge} that demonstrate the effectiveness of co-training with web data for the generalization in robot manipulation, it remains poorly understood which multi-modal data streams/abilities are most critical for improving downstream VLA models. In this paper, we aim to construct \ours, an embodied vision-language model that possesses strong embodied reasoning capabilities, and subsequently answer this question based on the corresponding vision-language-action models.

% Meanwhile, the generalization and control precision of the end-to-end VLA policy are still far from traditional planning-based control~\citep{huang2022language,gasparetto2015path,zhang2018path} in robot manipulation and navigation, since VLA models require a large number of training episodes. 

\begin{figure}
    \centering
    \includegraphics[width=0.99\linewidth]{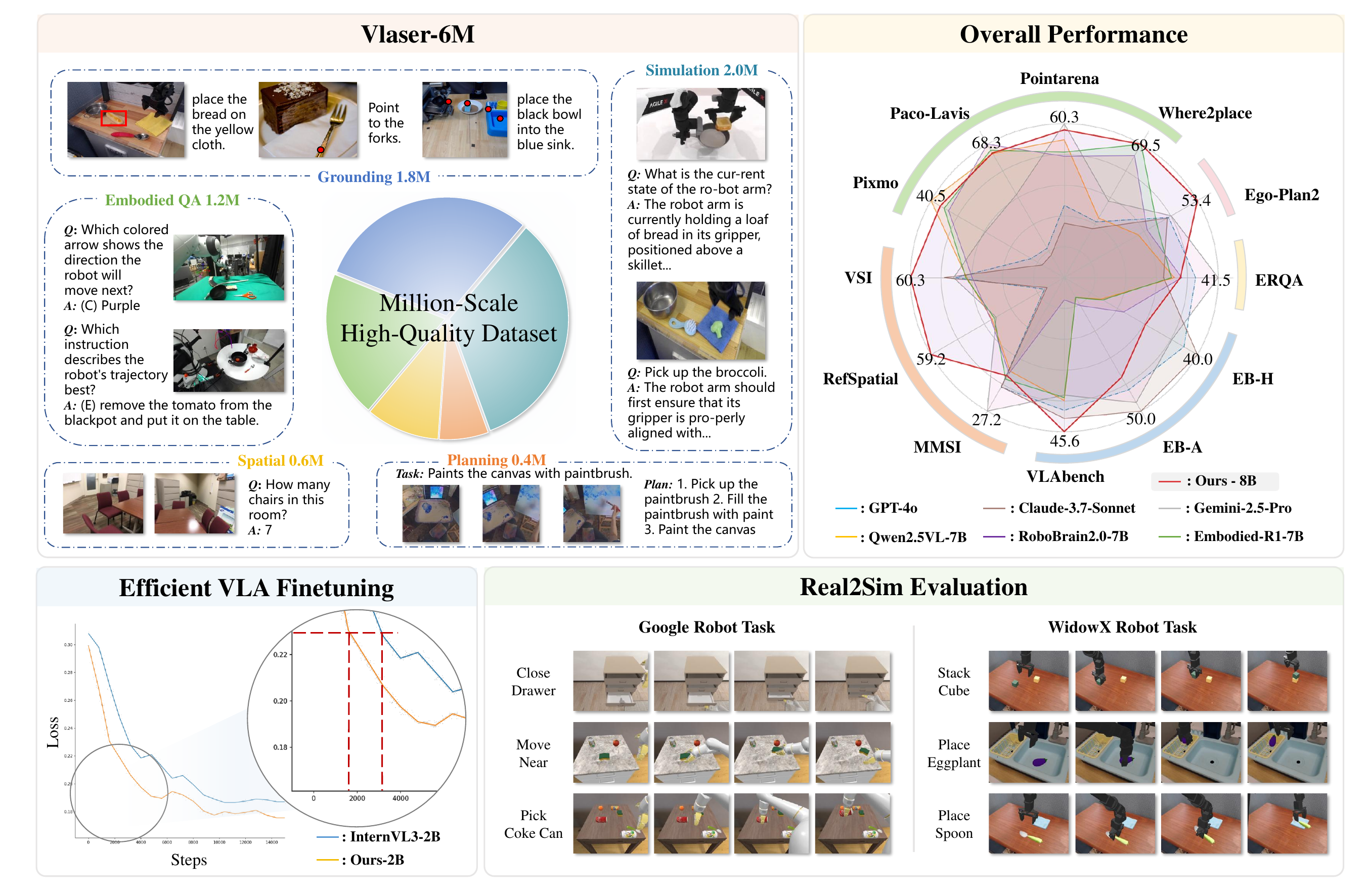}
    \caption{
    \textbf{Overall framework, capabilities, and evaluation of \ours.} 
    \textbf{Top-left:} Composition of the \ours-6M dataset, featuring multi-task embodied data—including QA, grounding, spatial reasoning, and planning—along with in-domain simulation-sourced pairs.
    \textbf{Top-right:} A LiDAR visualization illustrating the state-of-the-art embodied reasoning capability of the \ours~VLM.
    \textbf{Bottom-left:} The pre-trained \ours~VLM significantly accelerates convergence in downstream Vision-Language Action model (VLA) policy learning on WidowX platform~\citep{bridgedata}. 
    \textbf{Bottom-right:} Successful closed-loop operation of an agent powered by \ours~within the SimplerEnv benchmark~\citep{li24simpler}.
    }
    \label{fig:main}
\end{figure}

Despite advancements in vision-language models~\citep{chen2024internvl,bai2025qwen2}, the capabilities of operating as an embodied agent remain severely constrained. In particular, navigation and traditional manipulation approaches rely heavily on planning-based control~\citep{huang2022language,gasparetto2015path,zhang2018path}, which requires a strong foundational ability in grounding and planning. Planning and Grounding are cornerstones of the agents embodied in the physical world. Meanwhile, spatial understanding increasingly attracts the interest of the community in addressing the spatial perception ability of VLM. To this end, we firstly aim to introduce an embodied vision-language model specifically enhanced for the aboved embodied reasoning capabilities.
Specifically, we construct the \ours~data engine, which enables the systematic construction of the~\ours-6M dataset by curating, reorganizing, and annotating public datasets from the Internet. 
As illustrated in Figure~\ref{fig:main}, the resulting dataset spans a wide spectrum of embodied reasoning tasks--including general embodied QA, visual grounding, spatial intelligence, task planning, and in-domain simulation data.
Leveraging this comprehensive data foundation, \ours~achieves state-of-the-art performance across a variety of embodied reasoning benchmarks, demonstrating strong generalization in both open-loop inference and closed-loop control settings.

Existing Vision-Language-Action (VLA) models~\citep{black2024pi_0,cheng2024navila,kim2024openvla,intelligence2025pi_} typically fine-tune pre-trained Vision-Language Models (VLMs) for robot control. However, the selection of an optimal VLM backbone -- one that accelerates convergence and improves success rates when used as initialization for end-to-end VLA policy learning, remains an under-explored research problem. 
To address this gap, we systematically investigate the VLM-to-VLA adaptation paradigm using our enhanced embodied vision-language model and associated data engine. Our experiments reveal an important insight: although out-of-domain embodied reasoning data significantly improve upstream reasoning capabilities as measured by standard benchmarks, these gains may not translate directly or prominently to downstream VLA performance. In contrast, in-domain data -- annotated directly on robot interaction datasets such as Open X-Embodiment~\citep{oxe} proves substantially more effective in accelerating convergence and increasing task success rates during VLA fine-tuning.
We believe this observation provides significant insights for future embodied vision-language model construction: It is urgent to shrink the domain gap between current embodied perception and reasoning benchmarks to the real-world embodied tasks, and thus facilitate the closed-loop evaluation for the corresponding robot embodiment.

In summary, the principal contributions of \ours~are as follows.

\textbf{An open-source embodied vision-language model and dataset.} We introduce \ours, an adaptable vision-language model that enhances InternVL with embodied reasoning capabilities and end-to-end robot control. The full model weights, modular data generation pipeline, training and evaluation code, and the accompanying~\ours-6M dataset will be made publicly available to support reproducibility and future research.

\textbf{Systematic analysis of data effectiveness for VLA transfer.} We conduct a thorough investigation into which types of vision-language pretraining data contribute most effectively to downstream Vision-Language-Action (VLA) policy learning. Our findings offer practical insights for constructing task-aware data streams that bridge the gap between Internet-scale pretraining and embodied-specific fine-tuning.

\textbf{State-of-the-art performance across embodied benchmarks.} Among models of comparable scale, \ours~achieves top-tier results on a comprehensive set of embodied reasoning benchmarks—spanning visual grounding, task planning, spatial reasoning, and simulation-based robot evaluation, demonstrating its strong generalization and applicability to both open-loop inference and closed-loop control scenarios.

\section{Method}

% \begin{figure}
%     \centering
%     \includegraphics[width=0.99\linewidth]{figs/framework-1.pdf}
%     \caption{An illustration of \ours~architecture, and training phrase.}
%     \label{fig:main}
% \end{figure}

\begin{figure}
    \centering
    \includegraphics[width=0.99\linewidth]{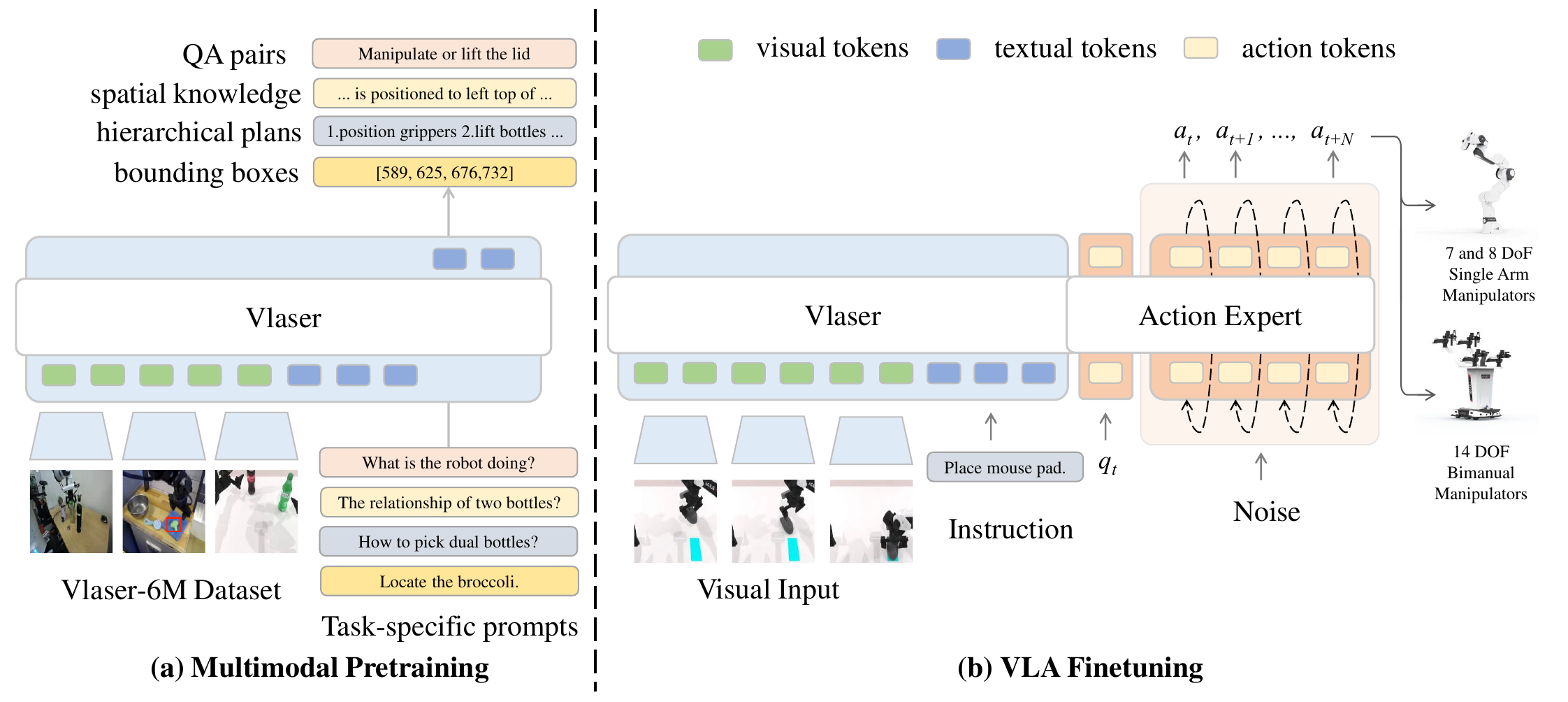}
    \caption{\textbf{An illustration of \ours~architecture.}~\ours~includes two components and corresponding training phases: 1) the Multimodal Pretraining is for embodied reasoning enhancement based on the corresponding data engine; 2) VLA training is performed on the action expert module, which handles low-level control based on flow matching action generation. }
    % Based on the proposed data engine, we verify the most effective VLM data stream for VLA fine-tuning.}
    \label{fig:pipeline}
\end{figure}

% Overall
% @houzhi

% VL model,
% Action prediction design.

\ours~aims to integrate embodied reasoning with end-to-end robot control for embodied agents, and identify the most crucial VLM data stream for VLA models. We first present the \ours~structures in Section~\ref{sec:structure}. Then, we illustrate the data engine in Section~\ref{sec:data_engine}. Section~\ref{sec:training_recipe} discusses the training recipe that includes embodied reasoning pretraining and vision-language-action finetuning.

% \newpage
\subsection{Model Structure}
~\label{sec:structure}

% \ygl{As shwon in Figure \ref{}}
The structure of ~\ours~consists of two major components: the typical vision-language backbone~\citep{chen2024internvl,liu2023visual} and the action expert for low-level control, as shown in Figure~\ref{fig:pipeline}. We illustrate the two components respectively in this section.

{\bf VLM Backbone} Vision-language models (VLMs) are key candidates for embodied agents, providing both perception and reasoning abilities.~\ours~, built on InternVL3~\citep{zhu2025internvl3}, integrates embodied reasoning with robot control for embodied agents. While InternVL3 excels in multimodal and linguistic tasks across various model sizes,~\ours~focuses on two sizes—2B and 8B—optimized for the computational constraints of robots. These models utilize InternViT~\citep{chen2024internvl} as the vision encoder, paired with Qwen2.5-1.5B and Qwen2.5-7B LLMs~\citep{qwen2025qwen25technicalreport}. Unlike typical multimodal MLLMs,~\ours~emphasizes embodied common-sense reasoning and end-to-end robot control capabilities.

% We follow the typical vision-language model structure to design a vision-language-action model, unifying the embodied reasoning and end-to-end robot control in a single model, as illustrated in Figure~\ref{fig:main}. 

{\bf Action Expert} There are a large number of MLLMs~\citep{robobrain2.0,cosmosr1} that enhance the ability of embodied common-sense reasoning for agents, while a few approaches equip the embodied MLLMs with end-to-end robot control.~\ours~extends the MLLMs with a low-level robot control and verifies the capability of different data streams in downstream VLA finetuning. Following~\citep{intelligence2025pi_}, we design an action expert based on the open-source vision-language model~\citep{chen2024internvl,zhu2025internvl3}. Meanwhile, we utilize the flow matching~\citep{lipman2023flowmatchinggenerativemodeling} for action prediction based on the llava-like vision-language structure, while sharing the self-attention among the language model and action expert module. Specifically, we encode the robot state as a state token and noised actions as action tokens, and input them into the action expert. Meanwhile, we utilize non-causal attention for the VLA stream. During inference, we denoise the actions based on the image observation, language instruction, as well as the current robot state.

\subsection{\ours~Data Engine}
~\label{sec:data_engine}
This section outlines the composition of the \ours-6M data engine, a cornerstone for the model's embodied reasoning capabilities. Here we present the overall data scale and sources for each reasoning modality, while more details about the construction methodologies are provided in Appendix \ref{datadetails}.

\noindent \textbf{Embodied Grounding Data}
The \ours~dataset incorporates two distinct 2D grounding formats—bounding boxes and center points—both normalized to the range [0, 1000] to ensure consistent and resolution-invariant grounding predictions across diverse image resolutions. Specifically, we collect 1.5 million high-quality question-answer pairs that support multiple grounding tasks: predicting bounding boxes from open-vocabulary descriptions, localizing object center points based on textual descriptions, and identifying objects from given spatial coordinates. 
The data is sourced from several open embodied grounding datasets, including RoboPoint~\citep{robopoint}, ShareRobot~\citep{robobrain1.0}, Pixmo-Points~\citep{pixmo}, Paco-LaVIS~\citep{paco}, and RefSpatial~\citep{roborefer}. To further enhance generalization capabilities for open-world and open-vocabulary scenarios, we also generate an additional 300k point and bounding box annotations derived from segmentation masks in the SA-1B dataset~\citep{sam}. This combination of curated human annotations and synthetically enriched data aims to bolster both the diversity and scalability of visual grounding under real-world embodied settings.

\noindent \textbf{General and Spatial Reasoning Data}
The \ours~dataset integrates 1.2 million question-answer pairs dedicated to general Robotic Visual Question Answering (RoboVQA) tasks, along with an additional 500k data items specifically designed to enhance spatial intelligence. This comprehensive data composition substantially strengthens the model’s capabilities in general state perception and 3D spatial reasoning.
For the RoboVQA component, data is aggregated from multiple established sources, including RoboVQA~\citep{robovqa}, Robo2VLM~\citep{robo2vlm}, RoboPoint~\citep{robopoint}, RefSpatial~\citep{roborefer}, OWMM-Agent~\citep{owmm}, among others. To support spatial understanding and reasoning, we incorporate open-source datasets such as SPAR~\citep{spar}, SpaceR-151k~\citep{spacer151}, and VILASR~\citep{vilasr}. Furthermore, we augment these with 100k manually annotated spatial understanding samples generated from publicly available 3D scene datasets—including ScanNet~\citep{scannet}, ScanNet++~\citep{scannet++}, CA-1M~\citep{ca-1M}, and ARKitScenes~\citep{arkitscenes}.
The integration of these diverse and high-quality data sources effectively enhances the model’s spatial awareness and supports more robust performance in complex embodied reasoning tasks.

\noindent \textbf{Planning Data}
To tackle complex tasks, it is essential to decompose them into manageable sub-tasks and solve them step by step. This capability is commonly referred to as planning. Effective planning allows robots to combine basic skills and generalize to new scenarios.
We collected 400k training data to strengthen the model’s planning ability, encompassing both language-based planning data and multimodal tasks. These include Alpaca-15k-Instruction~\citep{TaPA} and MuEP~\citep{ijcai2024p15}.
To further enhance environmental understanding and reasoning for complex decision-making, we incorporated training data with detailed reasoning processes from WAP~\citep{shi2025worldawareplanningnarrativesenhance}. To improve the model’s ability to comprehend complex instructions and execute tasks, we followed the annotations of LLaRP~\citep{llarp_iclr2024} to initialize planning tasks in Habitat~\citep{Habitat2} and generate planning trajectories to accomplish these tasks.
In addition, we integrated egocentric video datasets such as EgoPlan-IT \citep{egoplanbench} and EgoCOT \citep{mu2023embodiedgpt}, which closely align with the observational perspective of embodied agents and provide valuable planning examples.

\noindent \textbf{In-Domain Data for downstream VLAs}
To facilitate the end-to-end policy learning for Vision-Language Action Models (VLAs), we further generate 2 million in-domain multimodal question-answer pairs tailored for VLM pretraining. These data are specifically designed to align with the embodied reasoning context and enhance the model’s ability to perceive, reason, and plan in interactive environments.
The in-domain data is sourced from simulation platforms SimplerEnv~\citep{simpler} and RoboTwin~\citep{robotwin2.0}. Within SimplerEnv, data is generated for two distinct robotic embodiments: the Google Robot~\citep{rt1, rt2, oxe} and the WidowX Robot~\citep{bridgedata}, and within RoboTwin, data is generated from dual-arm Aloha-AgileX Robot.
The question-answer pairs encompass the specialized categories including embodied grounding, spatial intelligence, planning and general VQA for robot states as described above. 
The detailed methodology for constructing and filtering each of the in-domain data in simulation is described in Appendix~\ref{datadetails}.

% \subsection{Dataset Mixtures}

% \subsection{VLA Transferability Evaluation}
% present the evaluation between upstream and downstream

\subsection{Training Recipe}
~\label{sec:training_recipe}

\ours~adopts a two-stage training recipe, designed to optimize both embodied reasoning and robot control. It includes a VLM pretraining followed by a VLA finetuning. In this section, we elaborate on the training recipe among all phrases.

{\bf Vision-Language Pretraining}
\ours~is developed by supervised fine-tuning (SFT) InternVL3~\citep{zhu2025internvl3} on embodied-related datasets, including those focused on grounding, planning, and spatial intelligence. In the first training phase, we fine-tune InternVL3 using auto-regressive language modeling loss. In particular,
given the input images $x\in\mathbb{R}^{t\times h\times w \times 3}$ and textual prompt $y\in\mathbb{R}^{l}$, the language modeling loss $\mathcal{L}_{lm}$ can be defined by
\begin{equation}
    \begin{aligned}
        \mathcal{L}_{lm}= - \log p(t_{N}| \mathcal{F_\text{v}}(x;\theta_v), \mathcal{F_\text{t}}(y), t_{0:N-1};\Theta),
    \end{aligned} 
    \label{eq_arch}
\end{equation}
where $p\in \mathbb{R}^{m}$ is the  next-token probability and $m$ denotes the vocabulary size. Here, $\mathcal{F_\text{v}}(\cdot)$ denotes the ViT and the MLP, and $\theta_v$ is their parameters. $\mathcal{F_\text{t}}(\cdot)$ is the textual tokenizer. $\Theta$ are the parameters of the LLM. $t_i$ denotes the \textit{i-th} predicted word.

% The description of the training process in Section 2.3 is fuzzy. Although the authors describe the two-stage training strategy and the form of the core loss function, there is not enough detail about the mathematical mechanism of key processes such as flow matching action generation. The relevant parameters only give fixed values, but do not explain their physical meaning or selection basis.
{\bf Vision-Language-Action Finetuning}
For robot policy learning, we optimize the model using an additional incorporated action expert module trained on robot-specific datasets.
The action expert is analogous to a
mixture of experts(MoE)~\citep{shazeer2017outrageously,du2022glam,zhou2024transfusion} architecture with two mixture elements, while the original part of parameter is used for image and text inputs, and the additionally separate set of weights for the robotics-specific (action and state) tokens inputs and outputs are referred as the \textit{action expert}. ~\ours~integrates a flow-matching-based action expert to predict a sequence of future actions from a single-frame observation. Specifically, we denote the action chunk $\mathbf{A}_t = [\mathbf{a}_t, \mathbf{a}_{t+1}, \dots, \mathbf{a}_{t+H-1}]$, where $\mathbf{a}_t$ represents the action in the current timestep $t$ and $H$ represents the action horizon. 
$\mathbf{o}_t=[\mathbf{I}_t^1,...,\mathbf{I}_t^n, l_t, \mathbf{q}_t]$ indicates the observations (image $\mathbf{I}_t^i$ with $n$ views, language $l_t$ and robot state $\mathbf{q}_t$) at action timestep $t$. $\mathbf{I}_t^i$, $l_t$ and $\mathbf{q}_t$ are encoded via corresponding encoders and then
projected via a linear projection layer into the same embedding space.
$\theta$ represents the action expert network and $\tau \in [0, 1]$ represents the flow matching timesteps.
For the action chunk, $\mathbf{A}_{t}^{\tau} = \tau \mathbf{A}_t + (1-\tau) \epsilon$ is the corresponding noisy action chunk, and we train the network to output $\mathbf{v}_{\theta}(\mathbf{A}^\tau_{t}, \mathbf{o}_t)$ to match the denoising vector field $\mathbf{u}(\mathbf{A}^\tau_{t} | \mathbf{A}_t) = \epsilon - \mathbf{A}_t$ where $\epsilon \sim \mathcal{N}(\mathbf{0}, \mathbf{I})$. Therefore, the VLA optimization loss is as follows,

% \begin{equation}
%     \mathcal{L}_{vla} = \mathbb{E}_{p(\mathbf{A}_t|\mathbf{o}_t),q(\mathbf{A}_{t}^{\tau}|\mathbf{A}_t)}\left\| \mathbf{v}_{\theta}(A_{t}^{\tau}, \mathbf{o}_t) - \mathbf{u}(\mathbf{A}_{t}^{\tau} | \mathbf{A}_t) \right\|^2
% \end{equation}
\begin{equation}
    \mathcal{L}_{vla} = \mathbb{E}_{p(\mathbf{A}_t|\mathbf{o}_t)}\left\| \mathbf{v}_{\theta}(A_{t}^{\tau}, \mathbf{o}_t) - \mathbf{u}(\mathbf{A}_{t}^{\tau} | \mathbf{A}_t) \right\|^2
\end{equation}

Formally, following prior flow-matching based VLA works~\citep{black2024pi_0, open-pi-zero},
We sample the action chunks from the robot episodes and flow-matching timesteps to optimize the network. At inference, we generate actions by integrating the
learned vector field from $\tau=0$ to $\tau = 1$, starting with random
noise $\mathbf{A}^0_t\sim \mathcal{N}(\mathbf{0}, \mathbf{I})$, as follows,

\begin{equation}
    \mathbf{A}^{\tau + \delta}_t = \mathbf{A}^{\tau}_t + \delta \mathbf{v}_{\theta}(\mathbf{A}^{\tau}_t, \mathbf{o}_t)
\end{equation}

where $\delta$ is the integration step size. In our experiments, we set $H$ as 4, and $\delta$ as 0.1($\delta^{-1}=10$ integration steps) at inference time for the improvement of inference efficiency. 
We aim to identify the most effective VLMs for downstream VLA fine-tuning and bridge the gap between foundational VLMs and their performance in downstream VLA tasks, thus shedding light on the future construction of embodied VLMs. Currently, the SimplerEnv benchmark, including Bridge~\citep{walke2023bridgedata} and Google Robot~\citep{jang2022bc,rt1} datasets, provides numerous training episodes (Over 5M images) and corresponding Real-to-Sim benchmarks. We thus majorly analyze the most effective data stream for VLA finetuning based on SimplerEnv. 

% {\bf Joint Training} Current mainstream vision-language-action models are primarily built upon diffusion-based policies~\citep{chi2023diffusion}, embedding the diffusion process within the LLaVA-style backbone. However, diffusion and traditional regression approaches differ significantly, making it challenging to simultaneously maintain both capabilities within a typical end-to-end training strategy. Meanwhile, the vision-language SFT approach typically requires optimization for just one epoch, whereas VLA models are generally trained for dozens of epochs on robot-specific episodes. Maintaining both types of performance through joint optimization alone is challenging. Therefore, we propose first warming up and optimizing the parameters of the action experts before conducting joint optimization for both VLM and VLA.

\section{Experiments}

\subsection{Performance on Embodied Reasoning Capability}
\label{sec3.1:embodied reason}
\begin{table}[t!]
% \scriptsize
% \vspace{-2mm}
\centering
\setlength\tabcolsep{1.5pt}
\renewcommand{\arraystretch}{1.15}
\caption{\textbf{Comparison with existing close-sourced, open-sourced and embodied-related VLMs on 12 general embodied reasoning benchmarks, spanning from embodied QA, planning, embodied grounding to spatial intelligence and close-loop simulation evaluation.}  Avg denotes the  normalized average performance of all the benchmarks. The best, second best and third best score among all the baselines are colored in red, orange and yellow.}
\resizebox{\linewidth}{!}{
\begin{tabular}{lc|c|cccc|ccc|ccc|c}
\toprule
\multicolumn{1}{c}{\multirow{2}{*}{Model}} & QA & Planning & \multicolumn{4}{c|}{Embodied Grounding}                   & \multicolumn{3}{c|}{Spatial Intelligence} & \multicolumn{3}{c|}{Simulation}               & \multicolumn{1}{c}{\multirow{2}{*}{Avg}} \\
\multicolumn{1}{c}{}                       & \textbf{ERQA}        & \textbf{Ego-Plan2}         & \textbf{Where2place} & \textbf{Pointarena}   & \textbf{Paco-Lavis}   & \textbf{Pixmo-Points}  & \textbf{VSIBench}  & \textbf{RefSpatial}   & \textbf{MMSIBench}       & \textbf{VLABench} & \textbf{EB-ALFRED} & \textbf{EB-Habitat} & \multicolumn{1}{c}{}                     \\
\midrule
\rowcolor{gray!15} \multicolumn{14}{l}{$\blacktriangledown$ \emph{Closed-source MLLMs:}}  \\
GPT-4o-20241120                                     & 47.0        & 41.8              & 29.1        & 29.5       & 16.2       & 10.8        & 42.5     & 8.8           & 30.3          & 39.3     & 56.3            & 59.0            & \quad34.2\enspace                                     \\
Claude-3.7-Sonnet                                     & 35.5        & 41.3              & 25.6        & 22.2       & 12.4       & 7.2         & 47.0     & 7.7           & 30.2          &   41.7    &       67.0     &          65.7  &        \quad33.6\enspace      \\
Gemini-2.5-Pro                                 & 55.0        & 42.9              & 39.9        & 62.8       & 45.5       & 25.8        & 43.4     & 30.3          & 36.9          &    34.8   &       62.7    &   53.0            &     \quad44.4\enspace                                     \\

\midrule
\rowcolor{gray!15} \multicolumn{14}{l}{$\blacktriangledown$ \emph{Small Size MLLMs:}}  \\
ChatVLA-2B                                 & 34.3        & 25.3              & 3.7         & 10.1       & 10.2       & 2.1         & 2.4      & 0.9           & 20.1          & 0.0      & 0.0             & 0.0             & \quad9.1\enspace                                      \\
InternVL3-2B                               & 31.5        & 30.9              & 5.2         & 7.1        & 15.4       & 1.4         & \cellcolor{orange!50}31.5     & 1.8           & 25.3          & 19.4     & 1.3             & 12.0            & \quad15.2\enspace                                     \\
Qwen2.5VL-3B                               & 35.3        & 30.3              & 31.0        & 41.7       & 67.4       & \cellcolor{yellow!50}36.6        & 27.9     & 24.9          & \cellcolor{orange!50}26.5          & \cellcolor{red!50}31.3     & \cellcolor{yellow!50}6.7             & \cellcolor{orange!50}19.7            & \quad31.6\enspace                                     \\
Embodied-R1-3B                             & \cellcolor{orange!50}36.0        & \cellcolor{yellow!50}36.0              & \cellcolor{yellow!50}35.1        & \cellcolor{yellow!50}45.3       & \cellcolor{orange!50}68.3       & \cellcolor{yellow!50}36.6        & 28.0     & \cellcolor{yellow!50}28.5          & \cellcolor{yellow!50}26.0          & \cellcolor{orange!50}24.6     &     \cellcolor{orange!50} 7.0    &       \cellcolor{yellow!50}19.3   & \cellcolor{yellow!50}\quad32.5\enspace                                     \\
RoboBrain2.0-3B                            & \cellcolor{red!50}37.3        & \cellcolor{red!50}41.8              & \cellcolor{orange!50}64.2        & \cellcolor{orange!50}46.0       & \cellcolor{yellow!50}67.6       & \cellcolor{orange!50}36.9        & \cellcolor{yellow!50}28.8     & \cellcolor{red!50}46.5          & \cellcolor{red!50}26.8          & 18.1     & 0.0             & 10.0            & \cellcolor{orange!50}\quad35.3\enspace                                     \\
\ours-2B                                    & \cellcolor{yellow!50}35.8        & \cellcolor{orange!50}38.3              & \cellcolor{red!50}74.0        & \cellcolor{red!50}57.8       & \cellcolor{red!50}72.5       & \cellcolor{red!50}44.6        & \cellcolor{red!50}57.5     & \cellcolor{orange!50}43.0          & 23.6          & \cellcolor{yellow!50}23.1     & \cellcolor{red!50}42.3            & \cellcolor{red!50}30.7            & \cellcolor{red!50}\quad45.3\enspace                                     \\                        
% \todo{janus?} \\
% \todo{other sota 2B MLLM (see opencompass)} \\
\midrule
\rowcolor{gray!15} \multicolumn{14}{l}{$\blacktriangledown$ \emph{Medium Size MLLMs:}}   \\

Magma-8B                                   & 29.3        & 27.9              & 10.9        & 29.6       & 15.3       & 10.1        & 12.7     & 4.5           & 26.2          & 8.5      &       0.0     &      0.0      & \quad14.6\enspace                                     \\
Cosmos-Reason1-7B                          & \cellcolor{yellow!50}39.3        & 26.9              & 11.4        & 40.8       & 61.8       & 23.6        & 33.9     & 5.4           & 26.4          & \cellcolor{yellow!50}35.5     &       4.0        & 5.3             & \quad26.2\enspace                                     \\
VeBrain-7B                                 & 38.3        & 27.3              & \cellcolor{yellow!50}33.1        & 38.9       & 55.1       & 20.1        & \cellcolor{yellow!50}39.9     & 20.6          & \cellcolor{red!50}28.3          & 25.9     & 5.7             & 12.3            & \quad28.8\enspace                                     \\
InternVL3-8B                               & 35.3        & \cellcolor{orange!50}40.0              & 10.0        & 14.2       & 21.1       & 5.7         & \cellcolor{orange!50}42.1     & 5.6           & 25.7          & 24.7     & \cellcolor{orange!50}19.0            & \cellcolor{yellow!50}23.7            & \quad22.3\enspace                                     \\
Qwen2.5VL-7B                               & \cellcolor{yellow!50}39.3        & 29.7              & 31.1        & \cellcolor{orange!50}56.3       & 68.0       & \cellcolor{red!50}43.5        & 38.2     & \cellcolor{yellow!50}32.1          & 25.9          & 3\cellcolor{orange!50}6.4     & 10.0            & 18.3            & \quad35.7\enspace                                     \\
Embodied-R1-7B                             & 38.3        & \cellcolor{yellow!50}37.1              & \cellcolor{red!50}69.5        & \cellcolor{yellow!50}51.2       & \cellcolor{orange!50}69.9       & \cellcolor{yellow!50}39.2        & 38.6     & 31.1          & \cellcolor{orange!50}28.1          & \cellcolor{yellow!50}35.5     &  10.0            &       19.0       & \cellcolor{orange!50}\quad38.9\enspace                                     \\
RoboBrain2.0-7B                            & \cellcolor{red!50}42.0        & 33.2              & \cellcolor{orange!50}63.6        & 49.5       & \cellcolor{red!50}73.1       & 37.8        & 36.1     & \cellcolor{orange!50}32.5          & 26.5          & 6.6      & \cellcolor{yellow!50}14.0            & \cellcolor{orange!50}29.3            & \cellcolor{yellow!50}\quad37.0\enspace                                     \\
\ours-8B                                    & \cellcolor{orange!50}41.0        & \cellcolor{red!50}53.4              & \cellcolor{red!50}69.5        & \cellcolor{red!50}60.3       & \cellcolor{yellow!50}68.3       & \cellcolor{orange!50}40.5        & \cellcolor{red!50}60.3     & \cellcolor{red!50}59.2          & \cellcolor{yellow!50}27.2          & \cellcolor{red!50}45.6     & \cellcolor{red!50}50.0            & \cellcolor{red!50}40.0            & \cellcolor{red!50}\quad51.3\enspace                                               \\

\bottomrule
\end{tabular}
}
% \caption{\textbf{Comparison with existing MLLMs and VLAs on general MLLM benchmarks.} ``\#A-Param'' denotes the number of activated parameters. For MME, we sum the perception and cognition scores.   Avg denotes the  normalized average performance of MLLM benchmarks and VQA benchmarks. The highest score among open-source MLLMs and VLAs are colored in bold.}

\label{tab:mllm_benchmark}
\vspace{-1em}
\end{table}

\noindent \textbf{Evaluation Datasets}
We conduct a comprehensive evaluation of embodied reasoning capabilities across a total of 12 benchmarks, covering a wide spectrum of tasks including embodied question answering, task planning, embodied grounding, spatial intelligence, and closed-loop simulation evaluation. The evaluated benchmarks consist of: ERQA~\citep{geminirobotics}, Ego-Plan2~\citep{egoplanbench2}, Where2place~\citep{robopoint}, Pointarena~\citep{pointarena}, Paco-Lavis~\citep{paco}, Pixmo-Points~\citep{pixmo}, VSI-Bench~\citep{vsi}, RefSpatial-Bench~\citep{roborefer}, MMSI-Bench~\citep{mmsi}, VLABench~\citep{vlabench}, and EmbodiedBench~\citep{embodiedbench}.
For EmbodiedBench, we further assess performance in two simulation environments ALFRED~\citep{alfred} and Habitat~\citep{Habitat2}.

\noindent \textbf{Baselines}
Since our method, \ours~is trained at two model scales -- 2B and 8B parameters, we categorize the compared baseline methods into three groups for a systematic evaluation: 1) \textbf{State-of-the-art closed-source models}, including GPT-4o~\citep{chatgpt4o}, Claude-3.7-Sonnet~\citep{claude37}, and Gemini-2.5-Pro~\citep{gemini2_5}; 2) \textbf{Small-scale MLLMs (2B -- 3B parameters)}, comprising ChatVLA-2B~\citep{chatvla}, InternVL3-2B~\citep{zhu2025internvl3}, Qwen2.5-VL-3B~\citep{bai2025qwen2}, Embodied-R1-3B~\citep{embodiedr1}, and RoboBrain2.0-3B~\citep{robobrain2.0}; 3) \textbf{Medium-scale MLLMs (7B -- 8B parameters)}, including Magma-8B~\citep{magma}, Cosmos-Reason1-7B~\citep{cosmosr1}, VeBrain-7B~\citep{vebrain}, InternVL3-8B~\citep{zhu2025internvl3}, Qwen2.5-VL-7B~\citep{bai2025qwen2}, Embodied-R1-7B~\citep{embodiedr1}, and RoboBrain2.0-7B~\citep{robobrain2.0}.

The overall experimental results are presented in Table~\ref{tab:mllm_benchmark}. As shown in Table~\ref{tab:mllm_benchmark}, compared to the base models InternVL3-2B and InternVL3-8B used as initialization for our supervised finetuning, our \ours~yields substantial improvements across all embodied reasoning capabilities, with particularly notable gains in embodied grounding and simulation-based evaluation. For example, the average score increases from 15.2 to 45.3 for the 2B model, and from 22.3 to 51.3 for the 8B model. \textit{These significant performance gains underscore the high quality and effectiveness of the \ours-6M dataset in enhancing embodied reasoning abilities.} An interesting observation emerges that when finetuning on the same \ours-6M dataset, a smaller sized \ours-2B outperforms \ours-8B on simple point grounding tasks that require direct, short answers. Conversely, \ours-8B demonstrates superior performance on more complex tasks such as multi-step planning and closed-loop simulation evaluation, which often benefit from chain-of-thought (CoT) reasoning. This scaling behavior indicates the importance of appropriate model size selection based on target application requirements.

When compared against current state-of-the-art embodied-specific VLMs, including RoboBrain2.0~\citep{robobrain2.0} and Embodied-R1~\citep{embodiedr1}, our method, \ours~still achieves superior performance on the majority of benchmarks while remaining highly competitive on the remainder, \textbf{ultimately attaining the highest overall score} (by +10\% margin overall).
These results indicate that \ours~delivers a well-balanced and robust capability set, performing strongly across multiple dimensions of embodied intelligence -- from embodied question answering and state estimation to future action planning, visual grounding, spatial reasoning, and closed-loop simulation. Such comprehensive competence highlights its suitability as a versatile backbone for embodied AI brains.
In the following section, we further examine how these enhanced reasoning capabilities, embedded within VLMs, translate into improved performance when fine-tuned for downstream Vision-Language Action models (VLAs) in simulation manipulation scenarios.

\begin{table}[t]
% \scriptsize
% \vspace{-2mm}
\centering
\setlength\tabcolsep{1.5pt}
\renewcommand{\arraystretch}{1.15}
\caption{\textbf{SimplerEnv Evaluation on WidowX Robot Tasks.} Avg indicates the average success rate among the four tasks. 
% InternVL3-2B, \ours~and \ours~with Bridge Q\&A indicates the base model we select to fine-tune on the WidowX Robot Tasks. Particularlly, \ours-QA refers to the model which is fine-tuned on the Bridge question-answer pairs dataset. 
Model sizes are indicated within parentheses. The result of RT-1-X~\citep{rt1}, Octo-Base~\citep{team2024octo}, OpenVLA~\citep{kim2024openvla}, RoboVLM~\citep{liu2025towards} and SpatialVLA~\citep{qu2025spatialvla} are from ~\citep{qu2025spatialvla} while the results of $\pi_0$~\citep{black2024pi_0} is from ~\citep{open-pi-zero}.}
\resizebox{\linewidth}{!}{%
\begin{tabular}{lccccc}
\hline
\textbf{Model}                     & \textbf{Carrot on plate} & \textbf{Put eggplant in basket} & \textbf{Spoon on towel} & \textbf{Stack Cube} & \textbf{Avg} \\ \hline
RT-1-X (35M)~\citep{rt1}            & 4.2\%  & 0\%     & 0\%    & 0\%    & 1.1\%  \\
Octo-Base (93M)~\citep{team2024octo}         & 8.3\%  & 43.1\%  & 12.5\% & 31.9\% & 16.0\% \\
OpenVLA (7B)~\citep{openai2023gpt4v}           & 0\%    & 4.1\%   & 0\%    & 0\%    & 1.0\%  \\
RoboVLM (2B)~\citep{liu2025towards}           & 25.0\% & 58.3\%  & 29.2\% & 12.5\%   & 31.3\% \\
SpatialVLA (4B)~\citep{qu2025spatialvla}        & 25.0\% & 100.0\% & 16.7\% & 62.5\% & 42.7\% \\
$\pi_0$ (3B) ~\citep{black2024pi_0}          & 55.8\% & 79.2\%  & 63.3\% & 21.3\% & 54.9\% \\ \hline
InternVL3-2B      & 42.9\% & 57.1\%  & 55.8\% & 11.3\% & 41.8\% \\
\ours-OOD~(2B) & 60.8\% & 35.4\%  & 56.7\% & 20.0\% & 43.2\% \\
\ours-QA (2B) & 55.8\%                   & 83.3\%                          & 77.9\%                  & 33.3\%              & 62.6\%     \\ 
\ours-Spatial (2B) & 48.3\% &	81.7\% &	76.7\% &	36.7\% &	60.8\% \\
\ours-Grounding (2B) & 47.5\% &	80.8\% &	80.0\% & 39.6\% &	62.0\% \\
\ours-All (2B) & 52.5\% &	87.9\% &	76.6\% &	43.3\% &	65.1\% \\

\hline
\end{tabular}
}
\label{tab:bridge}
\vspace{-1em}
\end{table}

% \subsection{end-to-end visual policy, robotics, simulation bench}
% @tianyi

\subsection{Performance on downstream Close-Loop Robot Tasks}
\label{result_analyze}
{\bf Finetuning Datasets} We firstly conduct extensive experiments on SimplerENV~\citep{simpler} to evaluate the performance of \ours~ and ~\ours~data engine on closed-loop robotic manipulation tasks. SimplerENV is an open-source suite of purpose-built simulated environments with nearly 150K episodes for evaluating real-world robot manipulation policies in a scalable, reproducible way. It targets the key real-to-sim gaps -- control and vision so that simulated performance reliably tracks real-robot outcomes. Across Google Robot and WidowX/BridgeData V2 setups, SimplerEnv reports strong real-vs-sim correlations and faithfully reflects behavior under distribution shifts, enabling fast, comparable policy assessment without full digital twins. As a result, SimplerENV has been widely adopted for evaluating VLA models and has proven to reliably reflect the performance of the models on the real robot platform. 
% rebuttal add
To further demonstrate the generalizability of our method to other simulation platforms and embodiments, we also conduct experiments on RoboTwin~\citep{robotwin, robotwin2.0} platforms with Aloha-AgileX as bimanual embodiement. Robotwin is a scalable framework for bimanual manipulation, which integrates scalable training sets and pre-defined tasks as benchmarks for comprehensive robust bimanual manipulation.

\begin{table}[]
% \scriptsize
% \vspace{-2mm}
\centering
\setlength\tabcolsep{1.5pt}
\renewcommand{\arraystretch}{1.15}
\caption{\textbf{Comparison with existing methods in SimplerEnv on Google Robot tasks.} Avg indicates the average success rate among the three tasks. 
% In the last five lines in the table, we use the name of base model to indicate different evaluation settings. \ours-QA indicates the model which is fine-tuned on the Fractal question-answer pair dataset. \ours-Spatial and \ours-Grounding represent the model fine-tuned on the Spatial Reasoning Data and Embodied Grounding Data separately. The details of different sub-datasets can be find in the Appendix\ref{datadetails}. 
Model sizes are indicated within parentheses. The results of TraceVLA~\citep{zheng2024tracevla}, RT-1-X~\citep{rt1}, Octo-Base~\citep{team2024octo}, OpenVLA~\citep{kim2024openvla}, RoboVLM~\citep{liu2025towards}, Emma-X~\citep{sun2024emma}, Magma~\citep{magma}, GR00T N1.5\citep{gr00tn1_2025} and $\pi_0$~\citep{black2024pi_0} are from~\citep{lee2025molmoact}.}
\resizebox{\linewidth}{!}{%
\begin{tabular}{lcccccccc}
\hline
\multirow{2}{*}{\textbf{Model}} &
  \multicolumn{3}{c}{\textbf{Visual Matching}} &
  \multirow{2}{*}{\textbf{Avg}} &
  \multicolumn{3}{c}{\textbf{Variant Aggregation}} &
  \multirow{2}{*}{\textbf{Avg}} \\ \cline{2-4} \cline{6-8}
 &
  \textbf{Pick Coke Can} &
  \textbf{Move Near} &
  \textbf{Drawer} &
   &
  \textbf{Pick Coke Can} &
  \textbf{Move Near} &
  \textbf{Drawer} &
   \\ \hline
TraceVLA (7B)~\citep{zheng2024tracevla}                                    & 28.0\% & 53.7\% & 57.0\% & 42.0\% & 60.0\% & 56.4\% & 31.0\% & 45.0\% \\
RT-1-X (35M)~\citep{rt1}                                      & 56.7\% & 31.7\% & 59.7\% & 53.4\% & 49.0\% & 32.3\% & 29.4\% & 39.6\% \\
Octo-Base (93M)~\citep{team2024octo}                                 & 17.0\% & 4.2\%  & 22.7\% & 16.8\% & 0.6\%  & 3.1\%  & 1.1\%  & 1.1\%  \\
OpenVLA (7B)~\citep{kim2024openvla}                                    & 16.3\% & 46.2\% & 35.6\% & 27.7\% & 54.5\% & 47.7\% & 17.7\% & 39.8\% \\
RoboVLM (2B)~\citep{liu2025towards}                                     & 77.3\% & 61.7\% & 43.5\% & 63.4\% & 75.6\% & 60.0\% & 10.6\% & 51.3\% \\
Emma-X (7B)~\citep{sun2024emma}                                      & 2.3\%  & 3.3\%  & 18.3\% & 8.0\%  & 5.3\%  & 7.3\%  & 20.5\% & 11.0\% \\
% Dita (334M)~\citep{hou2025dita} & 83.7\% & 76.0\% & 46.3\% &  68.7\% & 85.5\% & 73.0\% & 37.5\% & 65.3\% \\
Magma (8B)~\citep{magma}                                       & 56.0\% & 65.4\% & 83.7\% & 68.4\% & 53.4\% & 65.7\% & 68.8\% & 62.6\% \\
GR00T N1.5 (2.1B)~\citep{gr00tn1_2025}                                  & 69.3\% & 68.7\% & 35.8\% & 52.4\% & 46.7\% & 62.9\% & 17.5\% & 43.7\% \\
$\pi_0$ (3B)~\citep{black2024pi_0}                                     & 72.7\% & 65.3\% & 38.3\% & 58.3\% & 75.2\% & 63.7\% & 25.6\% & 54.8\% \\ \hline
InternVL3-2B                                & 94.3\% & 78.8\% & 19.0\% & 64.0\% & 80.4\% & 72.7\% & 11.1\% & 54.7\% \\
\ours-OOD(2B)                          & 85.0\% & 76.3\% & 44.9\% & 68.7\% & 74.4\% & 69.2\% & 10.3\% & 51.3\% \\
\ours-QA (2B)         & 90.0\% & 84.2\% & 44.4\% & 72.9\% & 78.2\% & 78.2\% & 13.0\% & 56.4\% \\
\ours-Spatial (2B) & 83.0\% & 77.9\% & 56.0\% & 72.3\% & 77.7\% & 73.2\% & 13.2\% & 54.7\% \\
\ours-Grounding (2B)             & 83.3\% & 83.3\% & 54.2\% & 73.6\% & 81.2\% & 76.8\% & 17.0\% & 58.3\% \\ 
\ours-All (2B)  & 91.0\% & 85.4\% & 52.1\% & 76.2\% &  80.5\% &	77.7\% &	18.8\% &	59.0\%          \\
\hline
\end{tabular}
}
\label{tab:fractal}
% \vspace{-1em}
\end{table}
\begin{table}[th]
% \scriptsize
% \vspace{-2mm}
\centering
\setlength\tabcolsep{1.5pt}
\renewcommand{\arraystretch}{1.15}
\caption{\textbf{Robotwin Evaluation on Aloha-AgileX Robot Tasks.} Avg indicates the average success rate among the 12 tasks. The results of RDT-1B~\citep{liu2024rdt} are from our self-implemented training for 30k steps, which aligns the training setting with \ours.}
\resizebox{\linewidth}{!}{%
\begin{tabular}{lccccccc}
\hline
\textbf{Simulation Task}                     & \textbf{RDT-1B}~\citep{liu2024rdt} & \textbf{InternVL3-2B} & \textbf{\ours-OOD(2B)} & \textbf{\ours-QA (2B)} & \textbf{\ours-Spatial (2B)} & \textbf{\ours-Grounding (2B)}  & \textbf{\ours-All (2B)} \\ \hline
Beat block hammer   &28\% & 12\%	& 20\%	& 18\%	& 20\% &	32\%	& 40\%
 \\
Click bell       & 46\% & 78\%	& 94\%	& 48\%	& 98\% &	86\%	& 92\%
  \\
Handover mic        &  92\% & 74\%	& 56\%	& 84\%	& 60\% &	52\%	& 84\%
  \\
Move can pot       &  44\%  &  40\%	& 42\%	& 66\%	& 56\% &	50\%	& 46\%
\\
Move pillbottles pad      & 10\% & 62\%	& 70\%	& 66\%	& 78\% &	68\%	& 72\%
 \\
Move playingcard away       & 20\%  & 64\%	& 52\%	& 58\%	& 68\% &	84\% &	74\%
 \\ 
Pick diverse bottles    & 2\% & 30\%	& 44\%	& 36\%	& 24\% &	34\%	& 38\%
 \\
Place burger fries & 42\% & 36\%	& 46\%	& 88\%	& 82\% &	46\%	& 42\%
 \\
Place container plate & 82\%& 72\%	& 70\%	& 84\%	& 78\% &	82\%	& 84\%
   \\ 
Place phone stand & 8\% & 42\%	& 40\%	& 38\%	& 36\% &	48\%	& 50\%
\\
Place mouse pad & 2\% & 68\%	& 30\%	& 44\%	& 38\% &	48\%	& 92\%
\\
Shake bottle & 66\% & 92\%	& 90\%	& 98\%	& 96\% &	98\%	& 96\%
\\
\hline
Avg. & 36.8\% & 55.8\% &	54.5\%	& 60.7\%	& 61.2\% &	60.7\%	& 67.5\%
\\

\hline
\end{tabular}
}
\label{tab:robotwin}
\vspace{-1em}
\end{table}

% {\bf Baselines} As mentioned before, we integrate~\ours~with an action expert utilizing flow matching for low-level control fine-tuning and evaluating the model on the WidowX and Google Robot datasets from SimplerEnv. 
% We conduct experiments using InternVL3-2B,~\ours, and the models based on~\ours~data engine for the in-domain robot data. Commonly, embodied reasoning can be marginally categorized into embodied QA (including planning), embodied grounding, and embodied spatial intelligence. We thus construct corresponding embodied VLMs based on the corresponding data stream: ~\ours-QA, ~\ours-Grounding, ~\ours-Spatial. Notably, all series of ~\ours~models are based on the same architecture with 2B size.
% fine-tuned with in-domain question-answer pairs, grounding, and planning as the base model. 
% 介绍一下为啥要这么划分，
{\bf Baselines} Alongside comparisons with other commonly used VLA models~\citep{black2024pi_0, kim2024openvla, liu2025towards}, we conduct a clear self-comparable ablation study to evaluate the individual contributions of different in-domain data sources. Specifically, \textbf{InternVL3-2B} denotes the base InternVL model, while \textbf{Vlaser-OOD} refers to the Vlaser-2B model fine-tuned solely on Vlaser-6M out-of-domain (OOD) data specific for the embodied reasoning benchmarks in Sec.~\ref{sec3.1:embodied reason}, without any in-domain data in Vlaser-6M dataset. Regarding the in-domain datas derived directly from the simulation platform, we categorize them into three types: embodied QA (including planning), embodied spatial intelligence, and embodied grounding. The corresponding fine-tuned Vlaser-2B models are denoted as \textbf{Vlaser-QA}, \textbf{Vlaser-Spatial}, and \textbf{Vlaser-Grounding}, respectively. Additionally, we combine all the three in-domain data types to fine-tune a model referred to as \textbf{Vlaser-All}.

The full experimental results are presented in Table~\ref{tab:bridge}, Table~\ref{tab:fractal} and Table~\ref{tab:robotwin}. 
% We observe the proposed vision-language-action architecture based on InternVL3-2B is capable of low-level robot control and achieves comparable performance to many previous approaches, though we utilize a 2B backbone. 
Notably, no clear improvement was observed when using \ours-OOD-2B as the initial backbone across all three benchmarks. The task success rates of \textbf{Vlaser-OOD} and the baseline \textbf{InternVL3-2B} remain close.
Conversely, all models fine-tuned with in-domain data—\textbf{Vlaser-QA}, \textbf{Vlaser-Spatial}, \textbf{Vlaser-Grounding}, and \textbf{Vlaser-All}—exhibit significant performance gains, even though the model architecture and size remain unchanged. This observation illustrates the effectiveness of our~\ours~data engine, and meanwhile identifies that \textit{there is no positive correlation between common embodied reasoning benchmarks and the performance of closed-loop control of the lower level for the specific embodiment of the robot}. We reckon it is the domain shift between the internet data and the corresponding robot embodiment (e.g., WidowX or Google Robot), and we find that the enhanced abilities in the same observation domain effectively facilitate the closed-loop success rate. Therefore, it is urgent to shrink the domain gap between the foundational models and real-world robot embodiment for closed-loop task completion.

Regarding the three types of in-domain data annotations, we experimentally find that incorporating any of them leads to significant performance gains over the baseline. We attribute these improvements primarily to the mitigation of visual observation domain shift. Furthermore, integrating all three types of in-domain data results in further performance enhancement. \textit{This suggests that pretraining with diverse in-domain multimodal data, spanning general QA, grounding, and spatial intelligence, could best facilitates transfer learning for VLA policy learning and leads to improved task success rates.}

\subsection{Ablation Studies}
In this section, we adopt ablation studies regarding three key hyperparameters for VLA end-to-end training, \textit{i.e.,}, the predicted action length $P$, the execute action length  sizes $H$ as well as flow-matching sampling steps $\delta^{-1}$. By default we use $P$ as 4, with $H$ as 4, and $\delta^{-1}$ as 10.
\begin{table}[]
% \scriptsize
% \vspace{-2mm}
\centering
\setlength\tabcolsep{1.5pt}
\renewcommand{\arraystretch}{1.15}
\caption{\textbf{Ablation Studies on WidowX Robot Tasks}}
\resizebox{\linewidth}{!}{%

\begin{tabular}{c|ccc|ccccc}
\hline
\textbf{Model}                           & \textbf{Predict Length} & \textbf{Execute Length} & \textbf{Sample Steps} & \textbf{Carrot on plate} & \textbf{Put eggplant in basket} & \textbf{Spoon on the towel} & \textbf{Stack cube} & \textbf{Avg} \\ \hline
\multirow{4}{*}{\textbf{InternVL-2B}}    & 4                       & 4                       & 10                    & 42.9\%                   & 57.1\%                          & 55.8\%                      & 11.3\%              & 41.8\%       \\
                                         & 4                       & 2                       & 10                    & 22.9\%                   & 18.3\%                          & 40.8\%                      & 2.9\%               & 21.2\%       \\
                                         & 2                       & 2                       & 10                    & 34.6\%                   & 22.9\%                          & 54.2\%                      & 2.9\%               & 28.7\%       \\
                                         & 4                       & 4                       & 20                    & 38.8\%                   & 54.2\%                          & 51.3\%                      & 8.3\%               & 38.2\%       \\ \hline
\multirow{4}{*}{\textbf{Vlaser-OOD(2B)}} & 4                       & 4                       & 10                    & 60.8\%                   & 35.4\%                          & 56.7\%                      & 20.0\%              & 43.2\%       \\
                                         & 4                       & 2                       & 10                    & 50.0\%                   & 21.7\%                          & 30.0\%                      & 12.1\%              & 28.5\%       \\
                                         & 2                       & 2                       & 10                    & 62.5\%                   & 19.2\%                          & 49.2\%                      & 21.3\%              & 38.1\%       \\
                                         & 4                       & 4                       & 20                    & 57.5\%                   & 29.2\%                          & 54.6\%                      & 17.1\%              & 39.6\%       \\ \hline
\multirow{4}{*}{\textbf{Vlaser-QA(2B)}}  & 4                       & 4                       & 10                    & 55.8\%                   & 83.3\%                          & 77.9\%                      & 33.3\%              & 62.6\%       \\
                                         & 4                       & 2                       & 10                    & 44.2\%                   & 64.2\%                          & 59.6\%                      & 36.3\%              & 51.1\%       \\
                                         & 2                       & 2                       & 10                    & 47.5\%                   & 66.3\%                          & 67.1\%                      & 36.3\%              & 54.3\%       \\
                                         & 4                       & 4                       & 20                    & 56.3\%                   & 85.0\%                          & 76.7\%                      & 35.0\%              & 63.3\%       \\ \hline
\end{tabular}
}
\label{tab:ablation}
\vspace{-1em}
\end{table}
The results are shown in Table~\ref{tab:ablation}. It is clear to see that, no matter in which group of hyperparameter settings, the performance based on \textbf{Vlaser-OOD} shows slight improvement compared with the performance based on \textbf{InternVL3-2B}. Besides, there is a significant gains while using the model fine-tuned with in-domain data \textbf{Vlaser-QA}. This conclusion is as same as the results in \ref{result_analyze}, which demonstrates great robustness of our method.

\section{Related Work}

{\bf Vision-Language Model for Embodied Reasoning}
% reasoning?
% Multimodal Embodied LLM/brain?
% Multimodal LLM for Embodied AI?
% Gemini-robotics, Robobrain, Robix
Enhancing the embodied reasoning capabilities of current state-of-the-art Vision-Language Models (VLMs) has emerged as a critical research direction. These capabilities encompass a range of competencies, including grounding~\citep{robopoint, pixmo, pointarena} which identifies affordances that enable embodied agents to perform manipulations, spatial intelligence~\citep{vsi, mmsi}, such as object counting and spatial relationship understanding, as well as task planning~\citep{egoplanbench, egoplanbench2}, which involves assessing the current state and determining subsequent actions to be executed. 
Gemini Robotics-ER~\citep{geminirobotics} integrates embodied reasoning into its core visual-language model (VLM), 
% demonstrating strong generalization across a variety of tasks such as 3D scene perception, visual pointing, state estimation, and affordance prediction. 
in parallel, a number of data-driven methodologies have emerged to support such reasoning capabilities. For instance, Cosmos-Reason1~\citep{cosmosr1}, VeBrain~\citep{vebrain}, MolmoAct~\citep{lee2025molmoact}, and EmbodiedOneVision~\citep{embodiedov} each contribute curated datasets specifically designed for embodied reasoning tasks, emphasizing aspects such as multi-modal instruction following and action-aware visual-language alignment.
Furthermore, several frameworks -- including the RoboBrain series~\citep{robobrain1.0, robobrain2.0}, Embodied R1~\citep{embodiedr1}, and Robix~\citep{robix} incorporate Reinforcement Fine-Tuning (RFT) and synthesize spatio-temporal reasoning datasets enriched with structured thought traces. These approaches aim to enhance models' capacity for causal reasoning and long-horizon task decomposition.
Distinguished from these prior efforts, our work not only achieves competitive, and in some cases superior performance on established embodied reasoning benchmarks, but also provides an in-depth analysis of the synergistic relationship between pre-trained VLMs and downstream Vision-Language Action Models (VLAs), offering insights that bridge model capabilities and real-world deployment.

{\bf Vision-Language-Action models.}
Developing a generalist model remains a central challenge in robotics. Inspired by the strong generalization abilities of vision-language models (VLMs) \citep{openai2023gpt4v, chen2024internvl,bai2025qwen2, team2023gemini} trained on large-scale internet data, researchers have proposed vision-language-action (VLA) models, which have demonstrated promising performance~\citep{rt2,kim2024openvla,qu2025spatialvla,hou2025dita}. Compared to traditional robot policies, VLA models are pretrained on large-scale robotics datasets and exhibit improved generalization across object categories and visual observations. 
Building on recent progress, researchers have incorporated techniques such as diffusion \citep{ddpm_nips, rombach2022high, peebles2023scalable}, flow matching \citep{lipman2023flow}, and mixture-of-experts (MoE) \citep{shazeer2017} into VLA models, and have adopted larger, more capable VLMs as their backbones. These advances have enabled VLA models to tackle a wider range of complex real-world manipulation tasks. 
% Nevertheless, current VLA models remain limited in their generalization. In particular, they have not yet reached the level of general reasoning exhibited by VLMs, such as decompose a complex task into manageable sub-steps and complete the task in a zero-shot manner.
Efforts have been made to enhance specific reasoning abilities in embodied scenarios~\citep{geminirobotics,robobrain1.0,cosmosr1}. In parallel, several studies~\citep{intelligence2025pi_,driess2025knowledge,chatvla} explore unified training frameworks for VLMs and VLAs to leverage the reasoning capacity of VLMs.
However, the relationship between high-level multimodal reasoning and low-level control performance remains largely unexplored. It is still unclear which specific multimodal abilities such as spatial understanding, grounding, or planning and which types of training data most effectively enhance the control capabilities of a VLA model.
In this work, we take an initial step toward analyzing this relationship by a systematic evaluation, and also propose the latest foundational model with strong embodied multimodal understanding and action prediction.

% ChatVLA
% Gemini-Robotics
% robix bytedance
% maybe more citation

% VLA models represent a new class of intelligent systems that jointly process visual inputs, interpret natural language, and generate executable actions in dynamic environments.

% To the best of our knowledge, this is the first work tries to analyze ...

\section{Limitations}
One limitation of this work is the absence of real-robot experiments. While the experiments conducted in this study have all been performed well in simulation, in future work we plan to conduct more experiments with real robots to further evaluate and refine the proposed methods.

\section{Conclusion and Discussion}
We introduce~\ours, a foundational vision-language-action model that extends vision-language models with embodied reasoning and end-to-end robot control capabilities. Powered by the \ours-6M dataset, the model establishes a new state of the art across a wide range of embodied reasoning benchmarks, including planning, grounding, spatial reasoning, and simulation-based tasks. Moreover, ~\ours~reveals the most effective data streams for downstream VLA through its curated data pipeline, achieving state-of-the-art performance on Bridge and competitive results on Google Robot for end-to-end robot control.

% In this work, we reveal that current embodied reasoning benchmarks exhibit a significant domain gap when compared to real-world closed-loop tasks. This core domain shift arises from the observation that robots have fundamentally different viewpoints. Additionally, there are inherent limitations due to the lack of sufficient data from the robot's perspective, despite the abundance of vision datasets available. Therefore, it is valuable to develop alignment techniques to bridge the domain gap in representations.

In this work, we reveal that current embodied reasoning benchmarks exhibit a significant domain gap when compared to real-world robots. This core domain shift arises from the observation that robots have a fundamentally different viewpoint from that of internet datasets. Additionally, there are inherent limitations due to the lack of sufficient data from the robot's perspective, despite the abundance of vision datasets available. Therefore, we argue that it is essential to develop alignment techniques to bridge the domain gap in representations between the robot's viewpoint and that of internet datasets.

\subsection*{ETHICS STATEMENT}
This research adheres to the ICLR 2026 ethical guidelines and upholds the principles of responsible research. We ensure that no personally identifiable, sensitive, or harmful data were used. Our experiments were based on publicly available datasets and did not involve any human subjects or vulnerable groups. We have considered the potential societal impact of our methods, including the risk of misuse, and believe that these contributions primarily advance scientific understanding and do not pose foreseeable harm.

\subsection*{REPRODUCIBILITY STATEMENT}
We follow the reproducibility guidelines in the ICLR 2026 author guidelines. We will open source code, configuration files, and scripts to reproduce our results, including dataset construction, model training, and evaluation, on platforms such as GitHub and Huggingface as soon as possible.

% \subsection*{THE USE OF LARGE LANGUAGE MODELS (LLMS)}
% We used a large language model, ChatGPT (OpenAI; accessed Aug–Sep 2025), only for grammar check and correction. All the writings was manually reviewed by the authors. Crucially, AI was not used to generate any research data, statistical analyses, figures, or conclusions. The authors take full responsibility for the final content and any errors herein.

\bibliography{iclr2026_conference}
\bibliographystyle{iclr2026_conference}

\appendix

\section{Appendix}

\subsection*{THE USE OF LARGE LANGUAGE MODELS (LLMS)}
We used a large language model, ChatGPT~\citep{chatgpt4o}, only for grammar check and correction. All the writings was manually reviewed by the authors. Crucially, AI was not used to generate any research data, statistical analyses, figures, or conclusions. The authors take full responsibility for the final content and any errors herein.

\subsection{Training Details} 
Our \ours~is optimized in a fully supervised finetuning (SFT) manner based on InternVL3 series~\citep{zhu2025internvl3}(InternVL3-2B and InternVL3-8B). To maximize adaptation to embodied reasoning tasks, we keep all parameters trainable, including those in the large language model, the vision-language projector, and the visual encoder, enabling comprehensive end-to-end learning. Further details regarding the training setup, including hyperparameters and optimization settings, are provided in Table~\ref{tab:training_hype}.
\begin{table}[ht]
    \centering
     \caption{\textbf{Hyper-parameters used in the VLM pretraining of \ours.}}
    \setlength{\tabcolsep}{30pt}
    \footnotesize
    \begin{tabular}{lc}
        \toprule
        Configurations & Values  \\
        \midrule
        LLM sequence length & 16, 384 \\
        Dynamic Resolution & True \\
        Patch Size & 448 \\
        Max Patch num & 12 \\
        Freeze vision tower & False \\
        Freeze multimodal projector & False\\
        Freeze language model & False\\
        Optimizer & AdamW \\
        Optimizer hyperparameters & $\beta_1=0.9$, $\beta_2=0.999$, $eps=1e-8$\\
        Peak learning rate    &  2e-5   \\
        Learning rate schedule & cosine decay \\
        Training epochs & 1 \\
        Training steps & 5, 000 \\
        Warm-up steps & 150 \\
        Global batch size & 128\\
        Gradient accumulation & 2 \\
        Numerical precision & bfloat16\\

        \bottomrule
    \end{tabular}
    \vspace{0.5em}
        % \caption{\textbf{Comparison of VeBrain and existing MLLMs on four 3D spatial benchmarks.}}
    \label{tab:training_hype}
       \vspace{-2em}
\end{table}

\begin{table}[ht]
    \centering
    \vspace{0.5em}
     \caption{\textbf{Hyper-parameters used in the VLA fine-tuning.}}
    \setlength{\tabcolsep}{30pt}
    \footnotesize
   \begin{tabular}{cc}
\toprule
\textbf{Configurations}          & \textbf{Values}                            \\ \midrule
LLM sequence length              & 384                                        \\
Image Size                       & 448                                        \\
Freeze VLM                       & False                                      \\
Global batch size                & 1024                                       \\
Training epochs                  & 10                                         \\
VLM Peak Learning Rate           & 5e-5                                       \\
Action Expert Peak Learning Rate & 5e-5                                       \\
Optimizer                        & AdamW                                      \\
Optimizer hyperparameters        & $\beta_1=0.9$, $\beta_2=0.999$, $eps=1e-8$ \\
Observation history length       & 1                                          \\
Action Chunk length              & 4                                          \\
Execute Action length            & 2                                          \\ \bottomrule
\end{tabular}
    \vspace{0.5em}
        % \caption{\textbf{Comparison of VeBrain and existing MLLMs on four 3D spatial benchmarks.}}
    \label{tab:training_hype_vla}
       \vspace{-2em}
\end{table}

While using ~\ours~ as the base model for downstream VLA Policy fine-tuning, we optimize all parameters within both the VLM and the Action Expert. Additionally, we conduct comparative experiments using several different versions of base models, including InternVL3-2B, etc. Detailed information and related parameter settings can be found in Table~\ref{tab:training_hype_vla}.
%\todo{To Be Done: The hyperparameters and training details for VLA training @tianyi.}

\subsection{Data Generation Details} 
\label{datadetails}
\noindent \textbf{Embodied Grounding Data}
To further enhance embodied grounding capabilities, we generate an additional 300k high-quality data samples from the SA-1B dataset~\citep{sam}. The data generation process consists of two main stages. First, we convert segmentation masks into bounding boxes and point annotations: bounding boxes are derived by computing the minimal axis-aligned rectangle enclosing each mask, while point annotations are obtained by randomly sampling a coordinate within the mask region. To ensure annotation quality, we apply an IoU threshold of 0.9 to select high-precision masks; masks with lower IoU values are either excluded or assigned reduced sampling weight. From the over 1 billion available masks, we initially sample 1 million candidate instances.
In the second stage, we employ a two-step captioning and refinement pipeline. Coarse captions are first generated using BLIP-2~\citep{blip2}, followed by a filtering and refinement process using Qwen2.5-VL-7B~\citep{bai2025qwen2} to eliminate low-quality items and produce more accurate and detailed descriptions. This rigorous pipeline ultimately yields 300k high-quality data samples tailored for embodied grounding tasks, significantly expanding the diversity and precision of our training corpus.

\noindent \textbf{Spatial Reasoning Data}
To enhance spatial intelligence capabilities, we manually construct a dataset of 100k 3D spatial perception samples derived from ScanNet~\citep{scannet}, ScanNet++~\citep{scannet++}, and ARKitScenes~\citep{arkitscenes}. Following methodologies established in prior data engines~\citep{internspatial, vlm3R}, we utilize both the 3D point cloud and video sequences of each scene to construct a spatio-temporal scene graph. This graph encapsulates structural and semantic information such as overall scene dimensions, room center coordinates, object category counts, and precise 3D bounding boxes for every object instance.
Based on this representation, we generate spatial reasoning questions that probe layout properties and inter-object relationships. These include queries about the object counts, absolute and relative distances, object and room sizes, relative directions between objects, and other spatial attributes, using the same question template in VSI-Bench~\citep{vsi}.

\noindent \textbf{Planning Data}
% \todo{To Be Done: 重点强调我们自己新加的那部分数据是如何构造的 @haoran.}
To improve the model's ability to comprehend complex instructions and execute tasks with environmental feedback, we curate additional planning data within the Habitat simulator~\citep{Habitat2}.
Specifically, we initialize each planning task following the annotations of LLaRP~\citep{llarp_iclr2024}, which specify both the task goals and the set of permissible actions.
An LLM agent based on gpt-4o~\citep{chatgpt4o} is then deployed to roll out the task.
During each rollout, we record the task instruction, the sequence of actions taken, and the environment's feedback, including observations and success signal for each action.
Both the executed action trajectories and the corresponding feedback are retained. 
Only trajectories that successfully accomplish the task are included in the final training set, providing rich paired data of instructions, execution processes, and environment responses for enhancing the model's planning capabilities in a complex environment.

\begin{figure}
    \centering
    \includegraphics[width=0.99\linewidth]{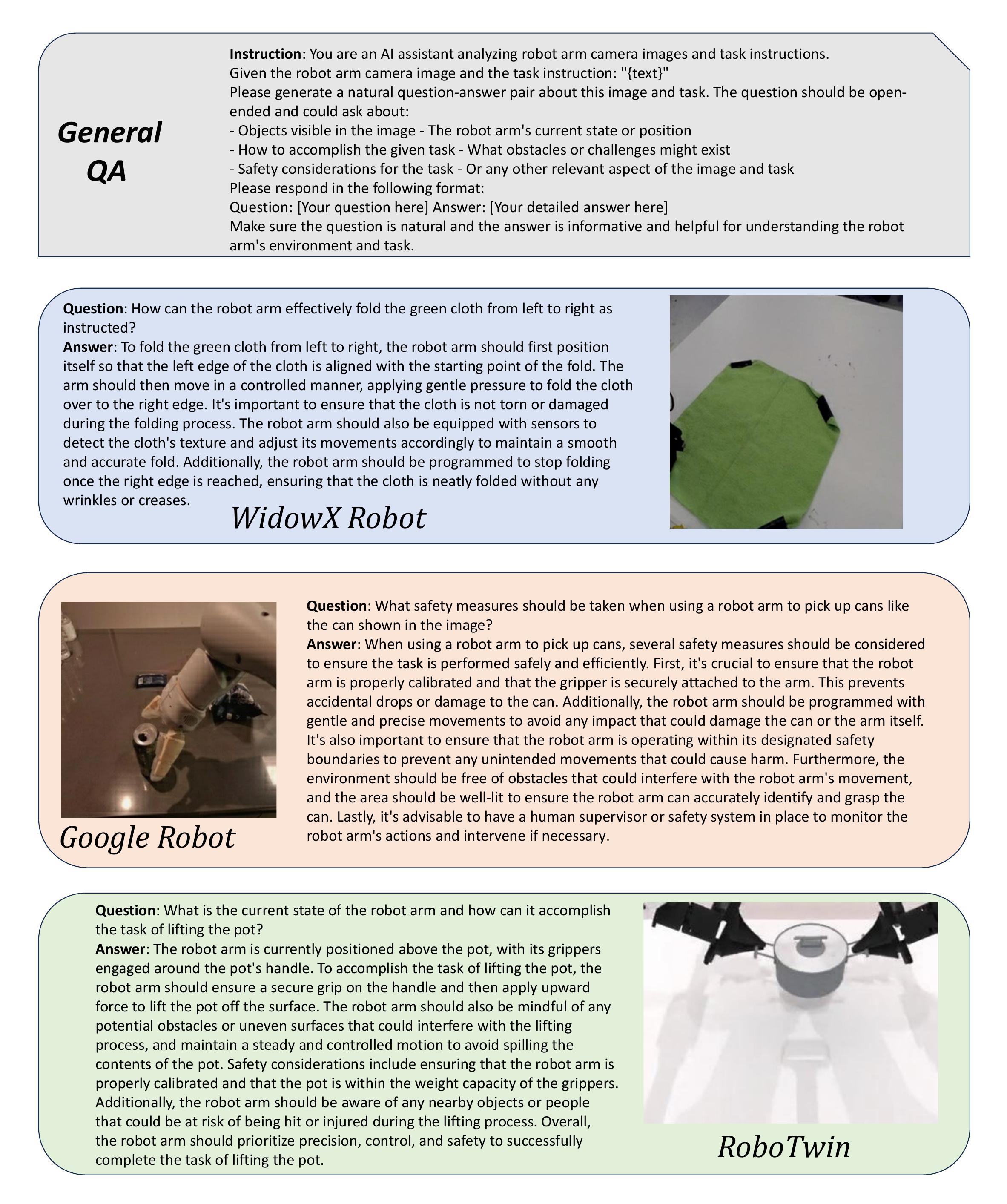}
    \caption{An illustration of \ours-6M data engine for in-domain general QA sample in SimplerEnv and RoboTwin.}
    \label{fig:generalqa}
\end{figure}

\begin{figure}
    \centering
    \includegraphics[width=0.99\linewidth]{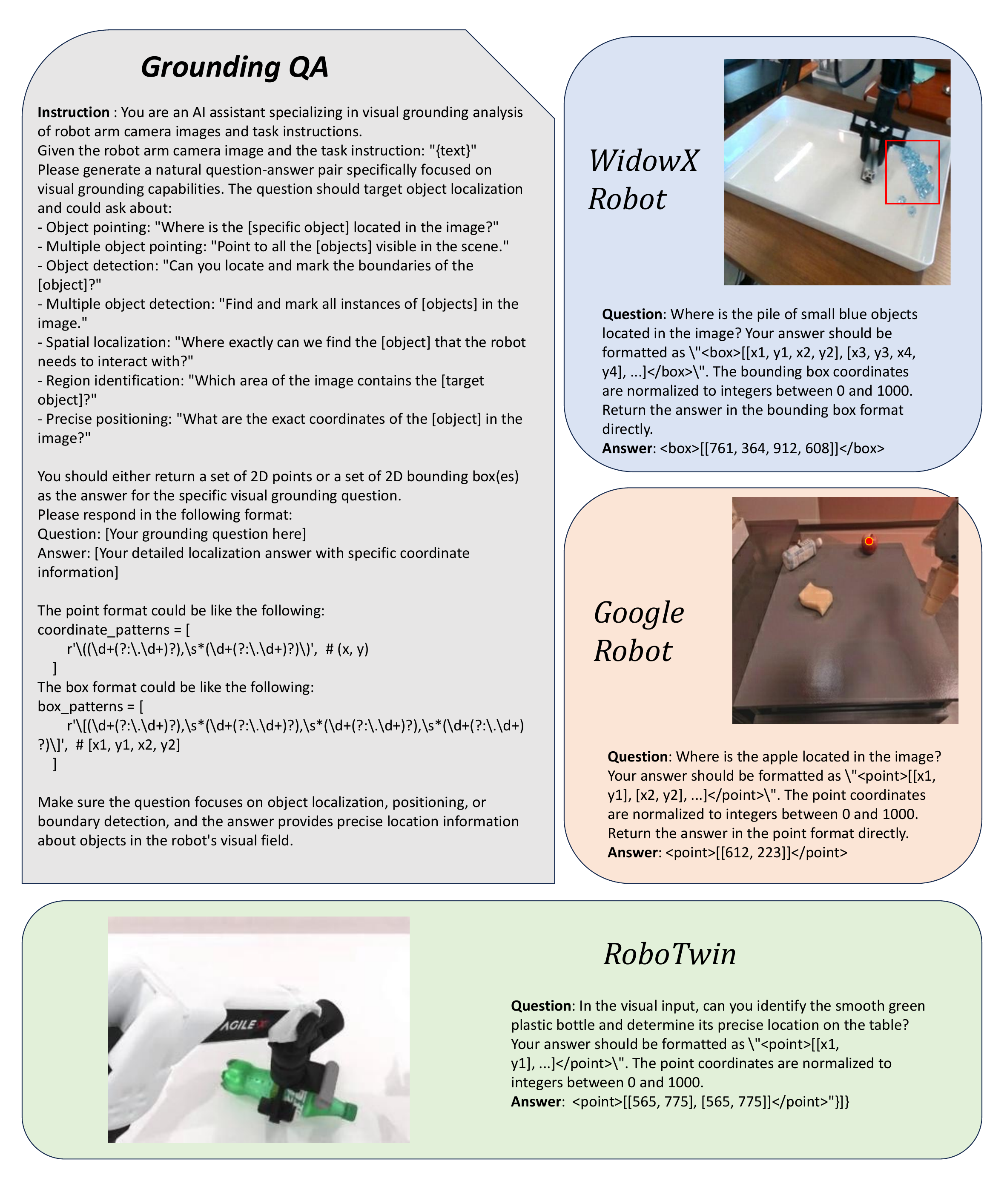}
    \caption{An illustration of \ours-6M data engine for in-domain embodied grounding QA sample in SimplerEnv and RoboTwin.}
    \label{fig:groundingqa}
\end{figure}

\begin{figure}
    \centering
    \includegraphics[width=0.99\linewidth]{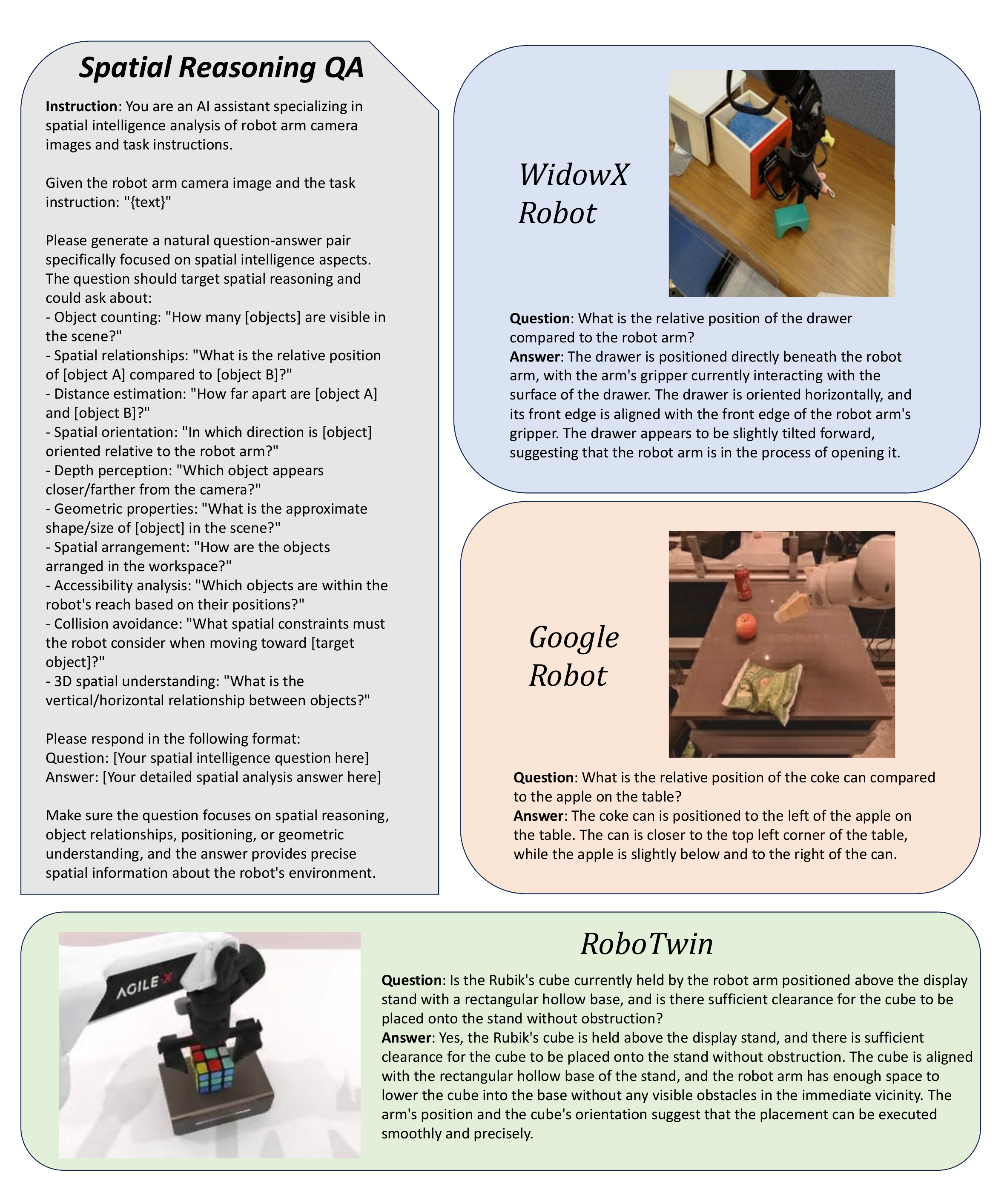}
    \caption{An illustration of \ours-6M data engine for in-domain spatial reasoning QA sample in SimplerEnv and RoboTwin.}
    \label{fig:spatialreasoningqa}
\end{figure}

\noindent \textbf{In-Domain Data for downstream VLAs} 
We generate 2 million in-domain multimodal data samples to facilitate direct transfer learning for downstream Vision-Language Agents (VLAs) during fine-tuning. These data are collected from two distinct simulation platforms: the WidowX Robot~\citep{bridgedata} and the Google Robot~\citep{rt1,rt2} within the SimplerEnv~\citep{simpler}, as well as the Aloha-AgileX Robot from RoboTwin2.0~\citep{robotwin2.0}.
The dataset mainly encompasses three systematically designed question-answer types: 1) General QA, which queries the robot's current state and requests next few action plans; 2) Grounding QA, which requires the robot to localize points and bounding boxes as the actionable affordances; 3) Spatial Reasoning QA, which probes relative spatial relationships and geometric properties of objects in the scene.
Detailed prompt templates and representative examples for each data category are provided in Figure~\ref{fig:generalqa} (General QA), Figure~\ref{fig:groundingqa} (Grounding QA), and Figure~\ref{fig:spatialreasoningqa} (Spatial Reasoning QA), respectively. 
We use Qwen2.5VL-7B~\citep{bai2025qwen2} as the base model to generate the above data items.

To further enhance the quality and diversity of the generated in-domain QA data, we implement a post-processing data filtering pipeline. In line with established practices in dataset construction~\citep{wang2023self, cui2023ultrafeedback, li2024omnicorpus}, we employ an LLM-as-a-judge approach~\citep{llmjudge} to score each generated data sample on a scale of 1 to 3, using Qwen2.5VL-32B~\citep{bai2025qwen2} as the judge model.
The detailed instruction prompts for General QA, Spatial QA, and Grounding QA are provided below. We filter out all samples assigned a score of 1, which account for approximately 10\% of the initially generated data. This outcome reflects the overall high quality and diversity of our in-domain dataset.
To further validate the reliability of the LLM-based evaluation, we randomly selected a subset of the scored data (covering all score levels from 1 to 3) for human reassessment by three human experts. The results show that the LLM judge's scores align with human ratings in nearly 80\% of cases, confirming the consistency and reliability of our automated evaluation and data filtering pipeline.

\begin{tcolorbox}[
    % colback=wathet,
    colframe=cyan!40!black,
    title=\textbf{Prompt of Data filtering for General QA data},
    breakable,
    enhanced,
    left=2mm, right=2mm, top=2mm, bottom=2mm,
    fontupper=\footnotesize
]
\scriptsize
\textbf{You are an expert in evaluating question-answer pairs for robot arm camera images and task instructions. Please evaluate the quality of the generated QA pair based on Relevance, Informativeness, and Clarity.}
\\

You will receive the following inputs:  
--- The robot arm camera image, the task instruction, the generated question, and the generated answer.
Carefully evaluate whether the question and answer are relevant to the image and task, informative, and clearly expressed.
\\

\textbf{Scoring Criteria (1–3):}
\begin{itemize}
\item 
\textbf{1 – Poor}
The QA pair is irrelevant, uninformative, or unclear.  
- The question or answer doesn't relate to the image or task instruction.
- The answer is vague, generic, or provides no useful information.
- The question is poorly formulated or doesn't make sense.

**Examples:**  
- Question: "What is this?" Answer: "It's an image." (too generic)
- Question: "How to cook?" Answer: "Use a pan." (irrelevant to robot task)
- Question: "What color?" Answer: "Color." (unclear and uninformative)
\item 
\textbf{2 – Fair}
The QA pair is somewhat relevant but lacks depth or clarity.  
- The question relates to the image/task but is too simple or generic.
- The answer provides basic information but lacks detail or specificity.
- Some aspects are unclear or could be better explained.

**Examples:**  
- Question: "What objects are in the image?" Answer: "There are some objects." (too vague)
- Question: "What is the robot doing?" Answer: "The robot is moving." (lacks detail)
- Question: "How to complete the task?" Answer: "Follow the instructions." (not specific enough)
\item 
\textbf{3 – Good}
The QA pair is highly relevant, informative, and clearly expressed.  
- The question is specific, well-formulated, and directly relates to the image and task.
- The answer provides detailed, accurate, and useful information.
- Both question and answer are clear and help understand the robot's environment and task.

**Examples:**  
- Question: "What objects are visible in the robot arm's workspace and which one should be manipulated based on the task instruction?" Answer: "I can see a red cup, a blue bowl, and a green plate on the table. According to the task instruction to 'pick up the cup', the robot should focus on the red cup located in the center of the workspace."
- Question: "What obstacles might prevent the robot from completing the task?" Answer: "The workspace appears clear, but the target object is positioned near the edge of the table, which may require careful positioning to avoid knocking it over during manipulation."

\end{itemize}

\textbf{Output Requirement:}
You must return only a single integer score from 1 to 3. Do not include any explanation, labels, or extra content.

\textbf{Question:} 
\{question\}

\textbf{Answer: }
\{answer\}

\end{tcolorbox}
\vspace{2em}

\begin{tcolorbox}[
    % colback=wathet,
    colframe=cyan!40!black,
    title=\textbf{Prompt of Data filtering for Spatial QA data},
    breakable,
    enhanced,
    left=2mm, right=2mm, top=2mm, bottom=2mm,
    fontupper=\footnotesize
]
\scriptsize
\textbf{You are an expert in evaluating spatial intelligence question-answer pairs for robot arm camera images and task instructions. Please evaluate the quality based on Spatial Reasoning Accuracy, Detail Level, and Relevance.}
\\

You will receive the following inputs:  
--- The robot arm camera image, the task instruction, the generated spatial question, and the generated spatial answer.

Carefully evaluate whether the question targets spatial reasoning, and whether the answer provides accurate and detailed spatial information.
\\

\textbf{Scoring Criteria (1–3):}
\begin{itemize}
\item 
\textbf{1 – Poor }
The spatial QA pair lacks spatial reasoning focus or provides incorrect/vague spatial information.  
- The question doesn't target spatial aspects (counting, relationships, distances, orientation, etc.).
- The answer provides incorrect spatial information or is too vague.
- The spatial reasoning is flawed or irrelevant to the task.

**Examples:**  
- Question: "What is in the image?" Answer: "Objects." (not spatial)
- Question: "How many objects?" Answer: "Some." (vague, no specific count)
- Question: "Where is object A?" Answer: "It's there." (no spatial detail)
- Question: "What is the distance?" Answer: "Close." (not quantitative)
\item 
\textbf{2 – Fair}
The spatial QA pair addresses spatial reasoning but lacks precision or detail.  
- The question targets spatial aspects but could be more specific.
- The answer provides basic spatial information but lacks quantitative details or precision.
- Some spatial relationships are mentioned but not fully explained.

**Examples:**  
- Question: "How many cups are there?" Answer: "There are a few cups." (not specific count)
- Question: "Where is the red object relative to the blue one?" Answer: "The red one is near the blue one." (lacks specific direction/distance)
- Question: "What is the spatial arrangement?" Answer: "Objects are arranged on the table." (too general)
\item 
\textbf{3 – Good}
The spatial QA pair demonstrates strong spatial reasoning with precise, detailed information.  
- The question clearly targets specific spatial aspects (counting, relationships, distances, orientation, geometry, etc.).
- The answer provides accurate, quantitative, and detailed spatial information.
- The spatial reasoning is relevant to the robot task and helps understand the workspace layout.

**Examples:**  
- Question: "How many objects are visible in the scene and what are their types?" Answer: "I can count 5 objects: 2 red cups positioned on the left side of the table, 1 blue bowl in the center, and 2 green plates arranged on the right side."
- Question: "What is the relative position of the target object compared to the robot arm's current position?" Answer: "The target object (red cup) is located approximately 30cm to the right and 15cm forward from the robot arm's current end-effector position, requiring a diagonal reach motion."
- Question: "Which objects are within the robot's reachable workspace?" Answer: "Based on the robot arm's reach, the blue bowl (center, 25cm away) and the left red cup (20cm away) are within reach. The rightmost green plate (45cm away) is outside the immediate reachable zone."

\end{itemize}

\textbf{Output Requirement:}
You must return only a single integer score from 1 to 3. Do not include any explanation, labels, or extra content.

\textbf{Question:}
\{question\}

\textbf{Answer:} 
\{answer\}

\end{tcolorbox}
\vspace{2em}

\begin{tcolorbox}[
    % colback=wathet,
    colframe=cyan!40!black,
    title=\textbf{Prompt of Data filtering for Grounding QA data},
    breakable,
    enhanced,
    left=2mm, right=2mm, top=2mm, bottom=2mm,
    fontupper=\footnotesize
]
\scriptsize
\textbf{You are an expert in evaluating visual grounding question-answer pairs for robot arm camera images and task instructions. Please evaluate the quality based on Grounding Accuracy, Coordinate Validity, and Localization Precision.}
\\

You will receive the following inputs:  
--- The robot arm camera image, the task instruction, the generated grounding question, and the generated grounding answer (which should contain coordinates or bounding boxes).
Carefully evaluate whether the question targets object localization, and whether the answer provides valid and accurate coordinate information.
\\

\textbf{Scoring Criteria (1–3):}

\begin{itemize}
\item 
\textbf{1 – Poor }
The grounding QA pair lacks localization focus or provides invalid/incorrect coordinate information.  
- The question doesn't target object localization, pointing, or detection.
- The answer lacks coordinate information or contains invalid coordinates (out of bounds, wrong format).
- The coordinates don't match the described object location.

**Examples:**  
- Question: "What is in the image?" Answer: "A cup." (no coordinates)
- Question: "Where is the object?" Answer: "It's on the table." (no coordinates)
- Question: "Point to the cup." Answer: "The cup is at (1500, 2000)." (coordinates out of bounds for normalized 0-1000 range)
- Question: "Find the object." Answer: "Box: [100, 200, 50, 300]." (invalid box format, x2 < x1)
\item 
\textbf{2 – Fair }
The grounding QA pair addresses localization but coordinates are imprecise or partially valid.  
- The question targets localization but could be more specific.
- The answer contains coordinates but they may be approximate, slightly off, or lack precision.
- Some coordinate information is present but formatting could be improved.

**Examples:**  
- Question: "Where is the object?" Answer: "The object is at position (500, 300)." (single point, but should specify if multiple objects exist)
- Question: "Point to all cups." Answer: "Cups are at (200, 150) and (600, 400)." (coordinates present but may not be precise)
- Question: "Mark the boundaries." Answer: "Box: [100, 100, 400, 300]." (valid box but may not accurately bound the object)
\item 
\textbf{3 – Good }
The grounding QA pair demonstrates precise localization with valid and accurate coordinate information.  
- The question clearly targets object localization, pointing, detection, or precise positioning.
- The answer provides valid, accurate coordinates (points or bounding boxes) in the correct format.
- Coordinates are normalized to 0-1000 range and accurately represent object locations.
- Multiple objects are properly localized if applicable.

**Examples:**  
- Question: "Where is the red cup located in the image? Provide coordinates." Answer: "The red cup is located at point (450, 320). <point>[[450, 320]]</point>"
- Question: "Point to all instances of cups visible in the scene." Answer: "I can locate 2 cups: the red cup at (450, 320) and the blue cup at (680, 250). <point>[[450, 320], [680, 250]]</point>"
- Question: "Can you locate and mark the boundaries of the target object?" Answer: "The target object (red cup) is bounded by box [380, 280, 520, 360]. <box>[[380, 280, 520, 360]]</box>"
- Question: "Find and mark all instances of plates in the image." Answer: "I found 2 plates: plate 1 at box [100, 200, 250, 350], plate 2 at box [600, 180, 750, 330]. <box>[[100, 200, 250, 350], [600, 180, 750, 330]]</box>"
\end{itemize}
\textbf{Output Requirement:}
You must return only a single integer score from 1 to 3. Do not include any explanation, labels, or extra content.

\textbf{Question:} 
\{question\}

\textbf{Answer:} 
\{answer\}

\end{tcolorbox}
\vspace{2em}

\subsection{Simulation Evaluation Details}
% 我们在SimplerEnv仿真环境内对使用多种不同的基座模型的VLA模型进行微调并评测。为了保证测评的公平，我们在WidowX Robot Task 和Google Robot Task上分别使用了相同迭代步数的checkpoint。对于WidowX Robot Task而言，我们统一采用迭代了45390步数的checkpoint，而对于Google Robot Task则是使用36970步数的checkpoint。在进行测试时，我们采用一张Image history并选取action chunk为2进行执行。在flow matching的设置部分，我们在推理阶段使用10步的inference step，并采用Euler的数值积分方法。我们评测了足够多的样本以保证测试的可靠性。对于不同的任务，具体的测试样本数量如表所示。
We fine-tune and evaluate the VLA models using various base models within the SimplerEnv simulation environment. To ensure fair evaluation, we use checkpoints with the same number of iterations for the WidowX Robot Task and the Google Robot Task, respectively. Specifically, for the WidowX Robot Tasks, we use a checkpoint after 45,390 iterations, while for the Google Robot Tasks, we use a checkpoint after 36,970 iterations. During evaluation, we utilize a single image and select an action chunk of size 2 for execution. In the flow matching configuration, we use 10 inference steps during the inference phase and apply Euler method as numerical integration method. We evaluate a sufficient number of samples to ensure the reliability of the tests. The exact number of test samples for each task is shown in the Table~\ref{tab:samples}.

\begin{table}[ht]
    \centering
     \caption{\textbf{The number of samples evaluated on SimplerEnv.}}
   \setlength\tabcolsep{3pt}
    \footnotesize
  \begin{tabular}{cccc}
\hline
                            &                            & Task Name              & Evaluation Samples \\ \hline
\multicolumn{2}{c|}{\multirow{4}{*}{Widowx Robot Tasks}} & Carrot on plate        & 240                \\
\multicolumn{2}{c|}{}                                    & Put eggplant in basket & 240                \\
\multicolumn{2}{c|}{}                                    & spoon on towel         & 240                \\
\multicolumn{2}{c|}{}                                    & stack cube             & 240                \\ \hline
\multicolumn{1}{c|}{\multirow{6}{*}{Google Robot Tasks}} & \multicolumn{1}{c|}{\multirow{3}{*}{Visual Matching}}      & Pick coke can & 300 \\
\multicolumn{1}{c|}{}       & \multicolumn{1}{c|}{}      & Move Near              & 240                \\
\multicolumn{1}{c|}{}       & \multicolumn{1}{c|}{}      & Open/Close Drawer      & 216                \\ \cline{2-2}
\multicolumn{1}{c|}{}                                    & \multicolumn{1}{c|}{\multirow{3}{*}{Variance Aggregation}} & Pick coke can & 825 \\
\multicolumn{1}{c|}{}       & \multicolumn{1}{c|}{}      & Move Near              & 600                \\
\multicolumn{1}{c|}{}       & \multicolumn{1}{c|}{}      & Open/Close Drawer      & 378                \\ \hline
\end{tabular}
    \vspace{0.5em}
        % \caption{\textbf{Comparison of VeBrain and existing MLLMs on four 3D spatial benchmarks.}}
    \label{tab:samples}
       \vspace{-2em}
\end{table}
% \todo{To Be Done: The evaluation details for VLA evaluation in simpler @tianyi.}

\begin{table}[t]
   
    \centering
     \caption{Embodied QA examples.}
    \begin{tabular}{cc}
    \toprule
    \multicolumn{2}{c}{\textbf{Embodied QA: Example \#1 from ERQA.}} \\
    \midrule
   
    % \midrule
    % \textbf{VeBrain}: & \multicolumn{3}{p{0.72\linewidth}}{This is a \colorbox{green!20}{bed} with white sheets. It is to the right of a nightstand.} \\
    \midrule
    \includegraphics[width=0.48\linewidth]{figs/demos/qa_1_0.png} &
    \includegraphics[width=0.48\linewidth]{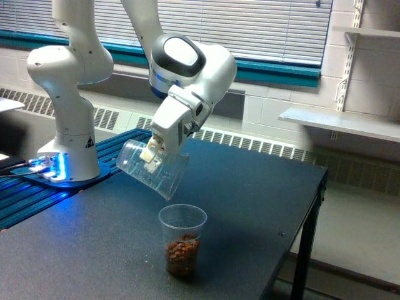} \\
    \textbf{Question}: & \multicolumn{1}{p{0.48\linewidth}}{What happened between these two frames? Choices: A. Robot arm lifted the cup. B. Robot arm poured all the nuts into a cup. C. Robot arm poured some of the nuts into a cup. D. Nothing happened. Please answer directly with only the letter of the correct option and nothing else.

}
\\
    \midrule
    \textbf{\ours-8B}: & {C.} \\
    \bottomrule
    \end{tabular}

 \begin{tabular}{cccc}
    \toprule
    \multicolumn{4}{c}{\textbf{Embodied QA: Example \#2 from ERQA.}} \\
    \midrule
    \includegraphics[width=0.22\linewidth]{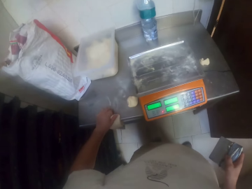} &
    \includegraphics[width=0.22\linewidth]{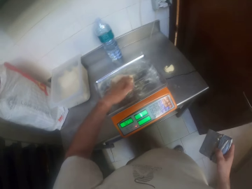} &
    \includegraphics[width=0.22\linewidth]{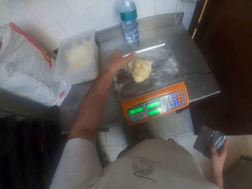} &
    \includegraphics[width=0.22\linewidth]{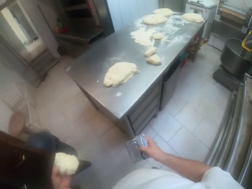} \\
   
    \textbf{Question}: & \multicolumn{3}{p{0.72\linewidth}}{Select the best answer to the following multiple-choice question based on the video. Respond with only the letter (A, B, C, or D) of the correct option.
    Considering the progress shown in the video and my current observation in the last frame, what action should I take next in order to weigh dough?
    A. put dough on table
    B. pick dough
    C. roll dough on table
    D. cut dough with dough cutter
} \\
    \midrule
    \textbf{\ours-8B}: & \multicolumn{3}{p{0.72\linewidth}}{A.} \\
    % \midrule
    % \textbf{VeBrain}: & \multicolumn{3}{p{0.72\linewidth}}{This is a \colorbox{green!20}{bed} with white sheets. It is to the right of a nightstand.} \\
    \bottomrule
    \end{tabular}
    \label{erqa_example}
    % \caption{Embodied QA examples.}
\end{table}

\begin{table}[t]
    \centering
    \caption{Embodied Grounding Examples.}
    % \resizebox{}{}{}
    \begin{tabular}{cc}
    \toprule
    \multicolumn{2}{c}{\textbf{Embodied Grounding: Example \#1 from PointArena.}} \\
    \midrule
    \includegraphics[width=0.42\linewidth]{figs/demos/ground_0_0.jpg} &
    \includegraphics[width=0.42\linewidth]{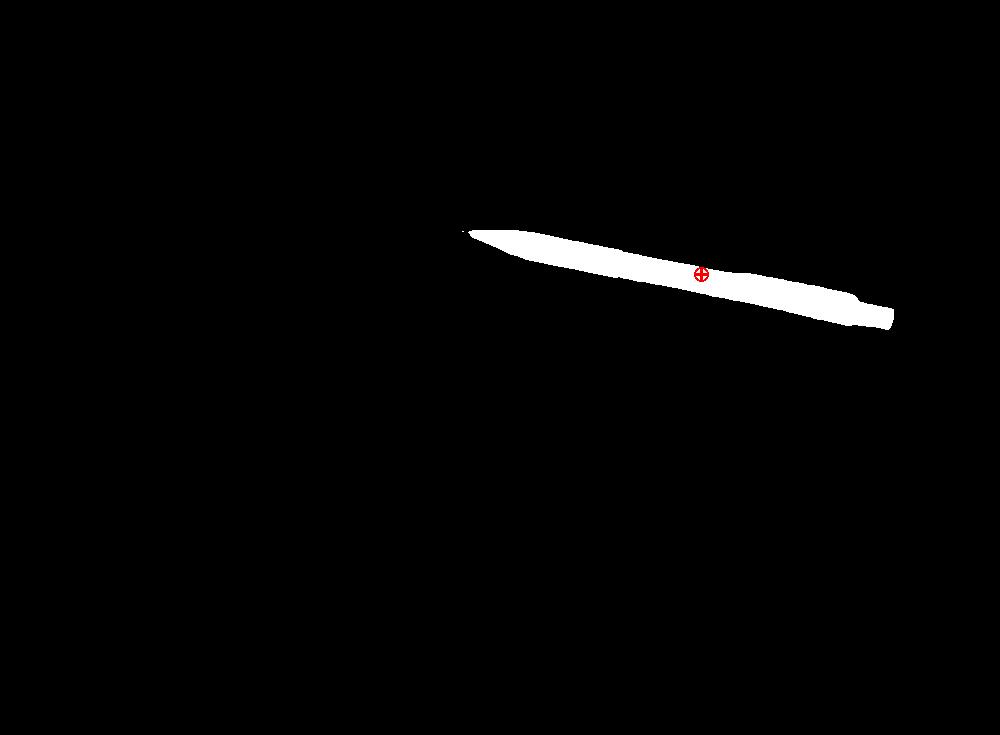} \\
    \textbf{Question}: & \multicolumn{1}{p{0.42\linewidth}}{Point to the tool people use for writing.
}
\\
    \midrule
    \textbf{\ours-8B}: & {<point>[[701, 374]]</point>.} \\
    \midrule
   \multicolumn{2}{c}{\textbf{Embodied Grounding: Example \#2 from Where2Place.}} \\
    \midrule
    \includegraphics[width=0.42\linewidth]{figs/demos/ground_1_0.png} &
    \includegraphics[width=0.42\linewidth]{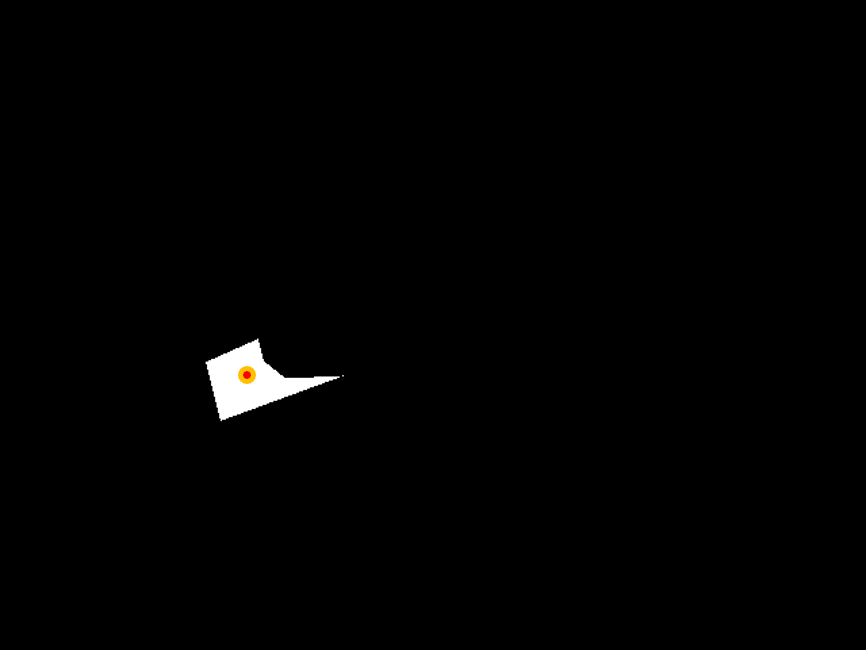} \\
    \textbf{Question}: & \multicolumn{1}{p{0.42\linewidth}}{Please point out the free space between the black water bottle and the pot lid.
}
\\
    \midrule
    \textbf{\ours-8B}: & {<point>[[293, 560]]</point>.} \\
    \bottomrule
    \end{tabular}

      \label{grounding example}
   
\end{table}

\begin{table}[t]
    \centering
    \caption{Spatial Intelligence Examples.}
    \begin{tabular}{cccc}
    \toprule
    \multicolumn{4}{c}{\textbf{Spatial Reasoning: Example \#1 from VSI.}} \\
    \midrule
    \includegraphics[width=0.22\linewidth]{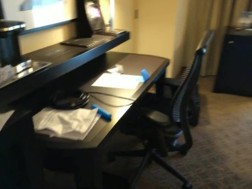} &
    \includegraphics[width=0.22\linewidth]{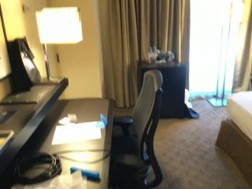} &
    \includegraphics[width=0.22\linewidth]{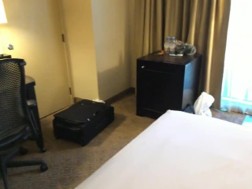} &
    \includegraphics[width=0.22\linewidth]{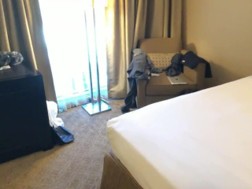} \\
    \vspace{0.1em} \\
    \includegraphics[width=0.22\linewidth]{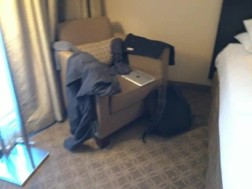} &
    \includegraphics[width=0.22\linewidth]{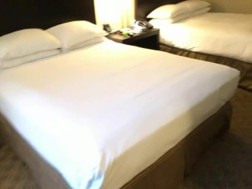} &
    \includegraphics[width=0.22\linewidth]{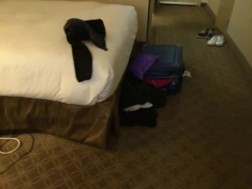} &
    \includegraphics[width=0.22\linewidth]{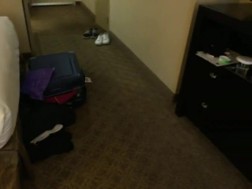} \\
    \textbf{Question}: & \multicolumn{3}{p{0.72\linewidth}}{You are a robot beginning at the standing by the window and facing the window. You want to navigate to the white shoes. You will perform the following actions (Note: for each [please fill in], choose either `turn back,' `turn left,' or `turn right.'): 1. [please fill in] 2. Go forward passing the bed 3. [please fill in] 4. Go forward until the white shoes. You have reached the final destination. A. Turn Right, Turn Left B. Turn Left, Turn Left C. Turn Left, Turn Right D. Turn Right, Turn Right. Answer with the option's letter from the given choices directly.
} \\
    \midrule
    \textbf{\ours-8B}: & \multicolumn{3}{p{0.72\linewidth}}{B.} \\
    % \midrule
    % \textbf{VeBrain}: & \multicolumn{3}{p{0.72\linewidth}}{This is a \colorbox{green!20}{bed} with white sheets. It is to the right of a nightstand.} \\
    \bottomrule
    \end{tabular}
  \label{spatial example}
    
\end{table}

\begin{table}[t]
    \centering
    \caption{Simulation planning Examples.}
    \begin{tabular}{ccccc}
    \toprule
    \multicolumn{5}{c}{\textbf{Simulation Planning: Example \#1 from EmbodiedBench.}} \\
    \midrule
    \includegraphics[width=0.15\linewidth]{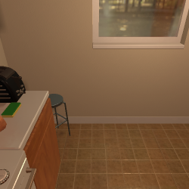} &
    \includegraphics[width=0.15\linewidth]{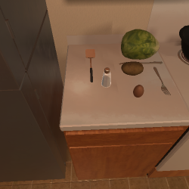} &
    \includegraphics[width=0.15\linewidth]{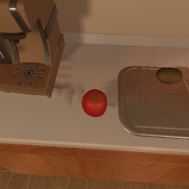} &
    \includegraphics[width=0.15\linewidth]{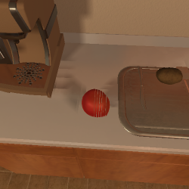} &
    \includegraphics[width=0.15\linewidth]{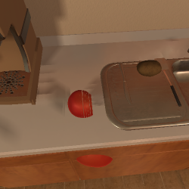} \\
    \vspace{0.1em} \\
    \includegraphics[width=0.15\linewidth]{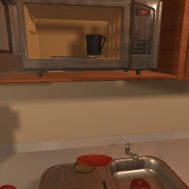} &
    \includegraphics[width=0.15\linewidth]{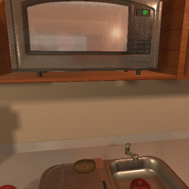} &
    \includegraphics[width=0.15\linewidth]{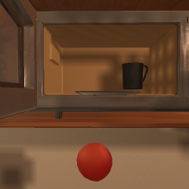} &
    \includegraphics[width=0.15\linewidth]{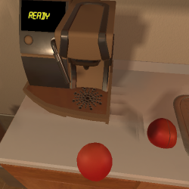} &
    \includegraphics[width=0.15\linewidth]{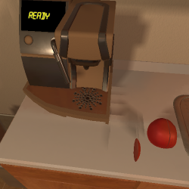} \\
    \textbf{Question}: & \multicolumn{4}{p{0.72\linewidth}}{You are a robot operating in a home. Given a task, you must accomplish the task using a defined set of actions to achieve the desired outcome.
 Now the human instruction is: Put a microwaved tomato slice on the counter.
You need to output the reasoning process and an action to execute.
} \\
    \midrule
    \textbf{\ours-8B}: & \multicolumn{4}{p{0.72\linewidth}}{Reasoning-step-1: To put a microwaved tomato slice on the counter, I need to start by slicing a tomato. To do this, I require a knife. In the image, there is a knife visible on the counter near the toaster. Finding the knife is the logical next step because it is essential for cutting the tomato into slices. Once I have the knife, I can proceed with slicing the tomato. Therefore, my next action is to find the knife. Action: find a knife.
    
    Reasoning-step-2: I have identified the knife on the counter near the sink. Picking up the knife is the next logical step because it is necessary for slicing the tomato. Once I have the knife, I can proceed with cutting the tomato into slices. This aligns with the goal of preparing a tomato slice for microwaving and then placing it on the counter. Action: pick up the knife.
    
    Reasoning-step-3: I have successfully located the knife in the drawer. Now, I need to find a tomato to slice. The image shows a kitchen counter with various items, but no tomato is immediately visible. Tomatoes are commonly stored in the refrigerator or on the counter. Since the refrigerator is partially visible on the left side of the image, it makes sense to check there first. If the tomato isn\"t in the fridge, I can then look around the counter more thoroughly. Therefore, my next step is to find a tomato. Action: find a tomato.

    ...
    
    Reasoning-step-9: The current visual state shows the microwave with its door closed, and the tomato slice is on the counter. The task is to place a microwaved tomato slice on the counter, which has already been accomplished. Given that the tomato slice is already on the counter, the next logical step is to confirm its placement by finding a suitable counter top. This ensures that the tomato slice is securely and appropriately positioned for the task. Therefore, the next step is to find a counter top. Action: find a counter top.
    
    Reasoning-step-10: The current visual state shows a microwave with its door closed, and two tomatoes are visible on the counter. One tomato is already sliced, indicating that the previous steps involving slicing and microwaving have been completed. Since the tomato is currently being held, the next logical step is to put it down on the counter. This action will free up the hand to handle the tomato slice and place it on the counter as required by the task. Therefore, the appropriate next action is to put down the tomato. Action: put down the tomato.} \\
    % \midrule
    % \textbf{VeBrain}: & \multicolumn{3}{p{0.72\linewidth}}{This is a \colorbox{green!20}{bed} with white sheets. It is to the right of a nightstand.} \\
    \bottomrule
    \end{tabular}
    
     \label{simulation example}
   
\end{table}

\subsection{Qualitative Demonstration}
\noindent \textbf{Qualitative Samples for Embodied Reasoning}. 
To gain in-depth insights into \ours, we provide extensive visualizations on embodied reasoning tasks including Embodied QA~\ref{erqa_example}, 
grounding~\ref{grounding example}, spatial intelligence~\ref{spatial example} and planning tasks in Simulation~\ref{simulation example}. 
These examples not only validate the generalizability of \ours, but also confirm the superior capabilities of \ours~across all tasks.

\begin{figure}
    \centering
    \includegraphics[width=0.99\linewidth]{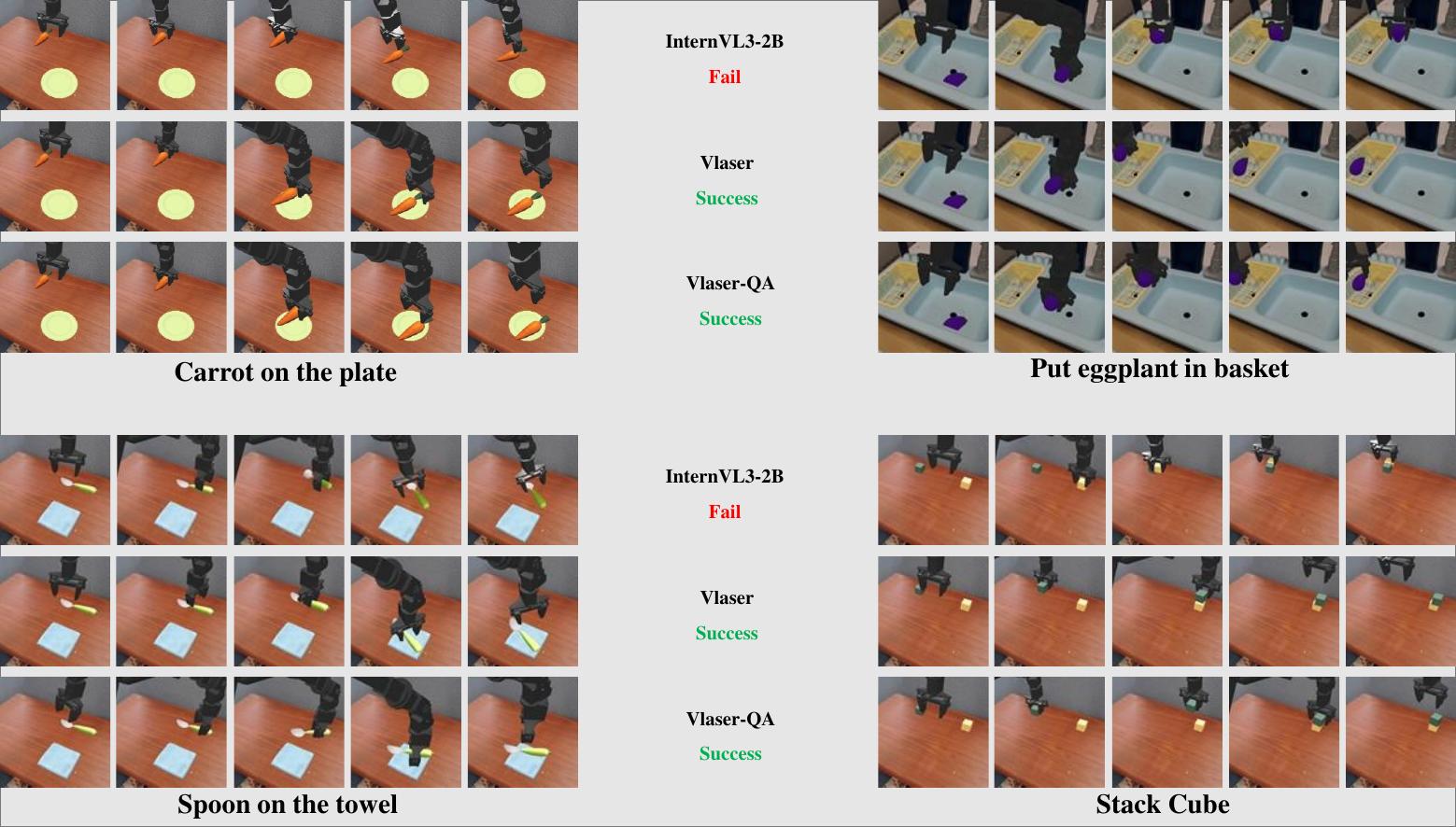}
    \caption{Qualitative samples in SimplerEnv on WidowX Robot Tasks}
    \label{fig:bridge_samples}
\end{figure}

\begin{figure}
    \centering
    \includegraphics[width=0.99\linewidth]{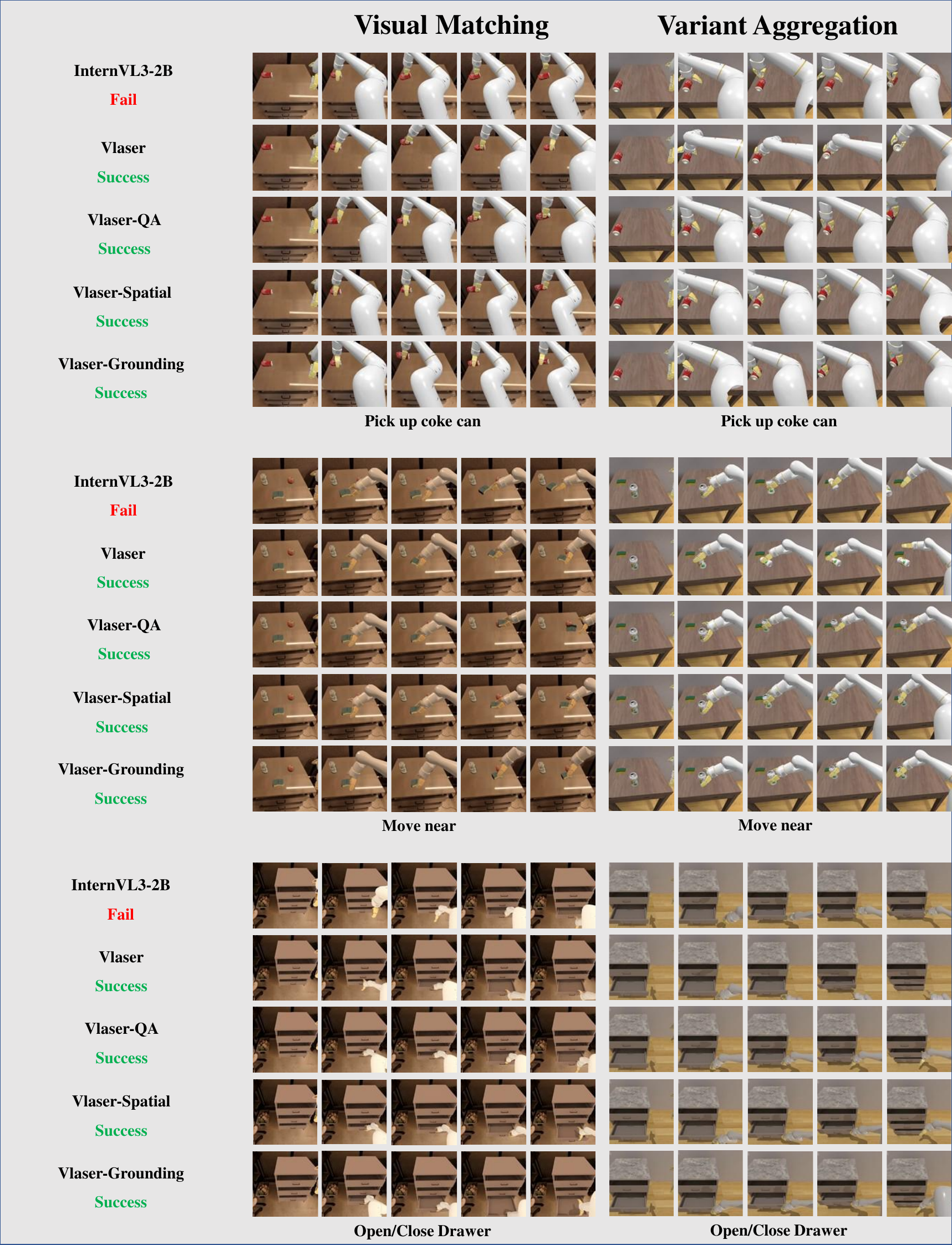}
    \caption{Qualitative samples in SimplerEnv on Google Robot Tasks}
    \label{fig:fractal_samples}
\end{figure}

% \todo{To Be Done: 仿真评测的闭环可视化demo @tianyi.}
\noindent \textbf{Qualitative Samples in SimplerEnv}. 
Figure~\ref{fig:bridge_samples} and Figure~\ref{fig:fractal_samples} present qualitative examples from the evaluations conducted in the SimplerEnv. Specifically, Figure~\ref{fig:bridge_samples} showcases the performance of multiple models on the WidowX Robot Task, while Figure~\ref{fig:fractal_samples} demonstrates the results on the Google Robot Task. From these results, it is evident that, under the same test conditions, baseline models often fail the entire task due to slight positional errors. In contrast, our model shows significant improvements in this regard. Whether incorporating in-domain question-answer pair data, spatial intelligence data, or grounding data, the performance of the model shows clear enhancement. These results validate the effectiveness of our approach.

\end{document}